\documentclass[journal]{IEEEtran}  

\IEEEoverridecommandlockouts                              

\usepackage{graphics} 
\usepackage{epsfig} 
\usepackage{mathptmx} 
\usepackage{times} 
\usepackage{amsmath} 
\usepackage{amssymb}  
\usepackage{cleveref}
\usepackage{algorithm}
\usepackage{algpseudocode}
\usepackage{xcolor}
\usepackage{csquotes}
\usepackage{booktabs}

\newcommand{\vect}[1]{\boldsymbol{#1}}

\title{\LARGE \bf Integrated Exploration-Aware \\UAV Route Optimization and Path Planning}

\author{Jimin Choi$^{1}$, Grant Stagg$^{2}$, Cameron K. Peterson$^{2}$, Max Z. Li$^{3}$
\thanks{*This work was supported by NSF IUCRC Phase I: Center for Autonomous Air Mobility and Sensing (CAAMS) Award No. 2137195.}
\thanks{$^{1}$Jimin Choi is with the Department of Aerospace Engineering, University of Michigan,
        Ann Arbor, MI 48109, USA; 
        {\tt\small jiminch@umich.edu}}%
\thanks{$^{2}$Grant Stagg and Cameron K. Peterson are with the Department of Electrical Engineering, Brigham Young University,
        Provo, UT 84602, USA; 
        {\tt\small \{ggs24, cammy.peterson\}@byu.edu}}%
        \thanks{$^{3}$Max Z. Li is with the Department of Aerospace Engineering, Department of Civil and Environmental Engineering, and Department of Industrial and Operations Engineering, University of Michigan, Ann Arbor, MI 48109, USA; 
        {\tt\small maxzli@umich.edu}}%
}

\begin{document}

\maketitle

\begin{abstract}
Uncrewed aerial vehicles (UAVs) are increasingly used for exploration-driven monitoring in hazardous environments such as disaster zones, contaminated sites, wildfire areas, and damaged infrastructure, where limited flight endurance must be allocated between visiting reported locations and gathering new information. In these settings, prior information regarding hazards is often incomplete, spatially imprecise, and subject to change during execution. For example, initial reports may identify a region where a hazard is likely to exist, but the actual hazard may be displaced, partially observed, or entirely unreported. We present an integrated exploration-aware UAV route optimization and path planning framework for hazard monitoring under uncertain and evolving prior information. The environment is represented as a spatial risk map, where each location has an associated belief of hazardous conditions. Reported hazards are modeled as uncertain regions of interest (ROIs) rather than confirmed target locations, requiring the UAV to inspect reported areas while also using its limited flight endurance to explore informative regions. The proposed method solves a vehicle routing problem over reported ROIs, augments the route with auxiliary pseudo-nodes to improve spatial coverage, allocates the remaining flight distance budget across route segments, and optimizes dynamically feasible B-spline trajectories for local exploration. During execution, UAV measurements update a grid-based belief map, and the remaining trajectory is replanned when new information and the remaining budget justify adaptation. Across 48 scenario configurations, online replanning improves average KL reduction by \(\mathbf{15.9}\%\) over the offline optimized planner and \(\mathbf{48.6}\%\) over straight-line traversal.
\end{abstract}


\section{Introduction}

Uncrewed aerial vehicles (UAVs) are increasingly used for exploration-driven monitoring missions in which limited flight endurance must be allocated between visiting reported locations and gathering new information. These missions arise in applications such as disaster response, industrial inspection, environmental monitoring, and hazard detection, where UAVs provide rapid, flexible, and remote assessment capabilities~\cite{MOHDDAUD202230, MISHRA20201}. Hazard monitoring is a particularly challenging real-world instance of this problem. In environments such as disaster zones, contaminated areas, wildfire regions, and damaged infrastructure, human responders face significant risks or access constraints, and autonomous agents must reason over uncertain and evolving information. Recent work has explored a wide range of UAV-assisted monitoring strategies, including routing optimization, cooperative UAV networks, and data-driven search and tracking methods~\cite{ERDELJ201772, wang_reinforcement_2019, choi2025bandit, choi2025airground}.

A central challenge in hazard monitoring is that prior information about hazards is often incomplete, spatially imprecise, and subject to change during execution. In practice, reports may be provided by human operators, external sensing systems, or initial reconnaissance, and these reports typically identify regions of interest (ROIs) rather than confirmed hazard locations. Such ROIs must be inspected as they represent potentially high-risk areas, but they do not fully characterize the environment. Hazards may be mislocalized, partially observed, entirely unreported, or newly revealed during execution, requiring UAVs to balance the inspection of reported ROIs with the exploration of surrounding uncertain regions. This setting is especially important in monitoring missions where mission-relevant risk information evolves over time. For example, changing wind conditions may reveal new wildfire hotspots, damaged infrastructure may expose secondary hazards such as gas leaks or unstable structures, and contamination plumes may shift as additional measurements are collected. In such cases, new sensor measurements can reveal discrepancies between the initial prior and the sensed environment, requiring the UAV to adapt its remaining plan. 

Many existing UAV routing approaches model the monitoring task as a vehicle routing problem (VRP) over a set of predefined targets~\cite{CALAMONERI2024123766}. While this formulation is useful for ensuring that known locations are visited under operational constraints, it is not sufficient when the environment contains spatially distributed risk and uncertain reports. Treating reported locations as deterministic hazard points can lead to overly rigid plans that focus only on nominal inspection targets and neglect informative exploration elsewhere. A more appropriate representation is a continuous risk field over the environment, in which ROIs define an imperfect prior belief and UAV measurements are used to refine that belief during execution.

Another limitation of many UAV monitoring methods is the separate treatment of routing and path planning. Routing determines the sequence in which ROIs or targets are visited, whereas path planning generates dynamically feasible trajectories between them. Most existing studies address these problems independently, and only a limited number of works have attempted to integrate them~\cite{Kiesel_Burns_Wilt_Ruml_2012, 9216819}. Even when routing and path planning are considered jointly, prior efforts often rely on predefined observation points or one-shot offline plans, leaving limited ability to adapt the path based on newly acquired risk information. This separation is problematic for hazard monitoring since the path flown between inspection locations can itself provide valuable sensing opportunities, especially when the initial ROI reports provide only an incomplete description of the risk field.

A prior conference paper~\cite{11312588} presented an earlier bi-level UAV monitoring framework that integrated route optimization with risk-aware B-spline path planning for hazard exploration. The present work makes three substantive extensions beyond that framework. First, it replaces discrete hazard targets with uncertain ROI reports over a continuous risk field, allowing reported regions to represent spatially imprecise and incomplete prior information rather than confirmed hazard locations. Second, it introduces streaming belief updates and online B-spline replanning during execution, so newly sensed or previously unreported hazards can influence the remaining trajectory. Third, it evaluates monitoring performance using risk field estimation metrics across hidden and mislocalized hazard scenarios. Together, these extensions yield an integrated framework that consistently accounts for the UAV's limited flight distance budget across route optimization and path-level exploration. It combines ROI-constrained routing, edge-based pseudo-node route augmentation, edge-wise allocation of the remaining flight distance, belief-weighted trajectory optimization, and online belief-driven replanning under imperfect and evolving information. Numerical experiments show that pseudo-node augmentation improves route-level spatial coverage, and across 48 scenario configurations, online replanning improves average Kullback-Leibler (KL) reduction by 15.9\% over the offline optimized planner and 48.6\% over straight-line traversal.

\section{Related Work}

\subsection{UAV Monitoring under Hazards and Uncertainty}

UAVs have been widely studied as autonomous monitoring platforms for disaster response, infrastructure inspection, environmental monitoring, and surveillance in hazardous environments~\cite{MOHDDAUD202230, MISHRA20201}. Prior work has considered emergency-response UAV platforms~\cite{brown2015sierra}, energy-constrained environmental monitoring~\cite{jensen2021nearoptimal}, cooperative sensing and disaster management~\cite{ERDELJ201772}, reinforcement learning for UAV monitoring under changing conditions~\cite{wang_reinforcement_2019}, and adaptive robust monitoring under model uncertainty~\cite{choi2026adaptive}. Surveys have also reviewed UAV monitoring and surveillance in uncertain, changing, or partially observed environments, including cooperative sensing, target search, target tracking, and networked surveillance~\cite{shakhatreh2019survey}. However, many existing approaches assume that targets, inspection locations, or monitoring regions are known in advance. This assumption is restrictive when reports are uncertain, spatially imprecise, or incomplete. Rather than treating prior reports as confirmed targets, our work models them as uncertain regions of interest and plans over a continuous risk field whose belief estimate is updated during execution.

\subsection{Route Optimization for UAV Monitoring Missions}

Route optimization is central to UAV monitoring since vehicles must visit multiple locations while satisfying flight-distance, battery, timing, and vehicle-availability constraints. Existing UAV routing formulations have incorporated endurance limits, refueling depots, multiple depots, heterogeneous vehicles, repeated trips, and revisit-period constraints~\cite{coutinho2018uavrouting}. Representative studies include UAV routing with refueling depots~\cite{sundar2014refueling}, cooperative routing for heterogeneous air-ground vehicle teams~\cite{manyam2020cooperative}, and multi-depot multi-trip UAV routing for post-disaster emergency response~\cite{CALAMONERI2024123766}. Routing has also been used for persistent surveillance, post-disaster inspection, and infrastructure damage assessment~\cite{stump2011persistent, chowdhury2021drone, lim2018multi}. However, most routing formulations assume a predefined set of locations to visit, which is limiting when hazard reports are uncertain or spatially imprecise and useful information may exist between reported regions. Our framework addresses this limitation by adding pseudo-nodes for broader spatial coverage and allocating marginal flight budget to route segments for downstream informative path planning.

\subsection{Informative Path Planning and Active Exploration}

Informative path planning and active sensing methods generate trajectories that maximize information gain, reduce uncertainty, or improve estimates of unknown spatial fields~\cite{popovic2024learning}. These methods have been applied to environmental monitoring, search and tracking, mapping, and robotic exploration under motion and budget constraints. Representative approaches include branch-and-bound informative planning~\cite{binney2012branch}, Bayesian optimization for informative continuous paths~\cite{marchant2014bayesian}, UAV terrain monitoring~\cite{popovic2020informative}, and adaptive informative path planning using deep reinforcement learning~\cite{ruckin2022adaptive}. However, informative planning methods are often studied separately from mission-level routing, with a fixed local region or prescribed start and goal locations. Our work connects mission-level routing and local informative planning by assigning edge-wise exploration budgets and updating paths online as the UAV's belief field changes.

\subsection{Integrated Routing and Path Planning}

A smaller body of work has studied the integration of routing and path planning, recognizing that high-level route decisions and low-level trajectory feasibility are coupled. For example, \cite{Kiesel_Burns_Wilt_Ruml_2012} studied integrated planning for route selection and motion feasibility, while recent work has considered combined routing and path planning formulations for uncrewed vehicles~\cite{9216819}. However, existing integrated routing and path planning approaches often remain centered on predefined targets, fixed observation points, or deterministic task specifications. As a result, they provide limited support for exploration-aware monitoring problems in which prior reports are uncertain, risk is spatially distributed, and newly collected measurements should influence the remaining trajectory. An earlier conference version of this framework combined route optimization with risk-aware B-spline path planning for UAV hazard exploration~\cite{11312588}. This paper is motivated by field settings such as wildfire assessment, disaster response, and contaminated site monitoring, where initial reports may only approximate the location and extent of hazardous conditions. We therefore extend prior framework to continuous risk field monitoring with imperfect ROI reports and online belief-driven replanning.

\section{Contributions of Work}

The contributions of this work are fourfold. First, we formulate UAV hazard monitoring as an uncertain ROI inspection and continuous risk field estimation problem, where reports are spatially imprecise and incomplete rather than confirmed target locations. Second, we develop a budget-consistent hierarchical planning framework that couples ROI-constrained routing with informative trajectory optimization through edge-wise exploration and pseudo-node route augmentation. Third, we propose a belief-driven B-spline planning and online replanning strategy that updates a grid-based log-odds belief field from streaming UAV measurements and adapts the remaining trajectory based on belief changes and residual budget. Fourth, we validate the framework across 48 scenario configurations with varying risk fields, hidden hazards, and pseudo-node settings. The proposed online replanning method is compared with three baselines. Offline optimized planning executes the initially planned belief-weighted trajectory without adaptation. Lawnmower coverage uses the available path budget for systematic geometric sweeping. Straight-line traversal directly connects route waypoints with minimal exploration. The results show that pseudo-node augmentation improves route coverage by \(91.6\%\) over ROI-only routing, and online replanning improves average KL reduction by \(15.9\%\) over offline optimized planning, \(36.3\%\) over lawnmower coverage, and \(48.6\%\) over straight-line traversal.

\section{Problem Statement}

We consider a UAV hazard monitoring mission over a bounded continuous spatial domain \(\Omega \subset \mathbb{R}^2\). The environment contains an unknown spatial risk field \(p_{\mathrm{true}}: \Omega \rightarrow [0,1]\), where \(p_{\mathrm{true}}(\vect{x})\) represents the true probability or intensity of hazardous conditions at location \(\vect{x} \in \Omega\). For example, \(p_{\mathrm{true}}(\vect{x})\) may represent the probability that a location contains wildfire activity, the normalized intensity of contamination, or the likelihood that damaged infrastructure poses a local safety risk. This field is used to support controlled simulation, measurement generation, and post-mission evaluation, but it is not known to the UAV during planning or execution. Instead, the UAV maintains and updates a belief field over the same domain. 
\begin{figure}[htbp]
    \centering
    \includegraphics[width=\linewidth]{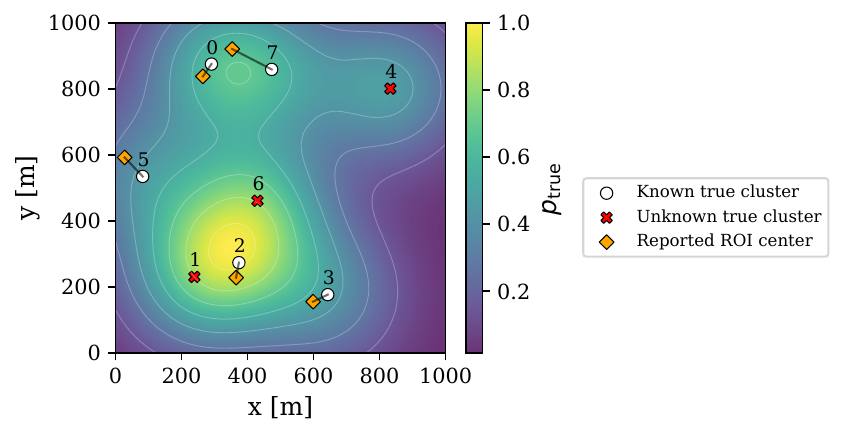}
    \caption{Example of a hazard monitoring environment over a continuous risk field. Reported ROI centers provide noisy and incomplete prior information. Some hazard clusters may be spatially mislocalized or unreported and must be discovered through exploration.}
    \label{fig:example_env}
\end{figure}

\Cref{fig:example_env} illustrates the risk field representation used in this work. The continuous risk field may be induced by multiple spatially localized hazard clusters. The UAV receives a set of reported ROI centers, shown as orange diamonds, which provide partial and noisy information about the environment. Some reports correspond to underlying hazard clusters, shown as white circles, but the reported centers may be spatially displaced from the true cluster locations. Other hazard clusters, shown as red crosses, may be unreported and can only influence the UAV's plan after they are sensed during execution.

The reported ROIs are therefore treated as uncertain inspection requirements rather than confirmed hazard locations. The UAV must inspect the reported ROIs while using its remaining flight budget to explore uncertain regions, collect measurements, and update its belief about the environment. This setting captures missions in which the initial reports provide only partial information and additional risk-relevant information may be revealed during execution. Previously unreported or newly sensed hazards can influence planning through the online measurement update and replanning mechanism.

Let the reported ROI set be denoted by \(\mathcal{R} = \{ r_1, r_2, \dots, r_{N_r} \}\), where each ROI \(r_i\) is represented by a center location \(\vect{c}_i \in \Omega\). Here, \(\vect{c}_i\) is a reported location in the same spatial domain as \(\vect{x}\), but it is not assumed to be the true maximizer or center of \(p_{\mathrm{true}}\). Instead, each ROI indicates a region where the prior belief of hazardous conditions should be elevated, while the true high-risk region may be displaced, broader, weaker, or absent. When available, each report may also include an uncertainty scale \(\sigma_i\) and a report strength \(w_i\), which describe the spatial precision and confidence of the ROI. The ROI reports are used to construct an initial belief field \(b_0: \Omega \rightarrow [0,1]\), which represents the UAV's prior estimate of the risk field before collecting new measurements. A typical construction combines a background risk level with local risk elevations around the reported ROIs:
\begin{equation}
    b_0(\vect{x}) =
    \sigma \left(
    \operatorname{logit}(p_{\mathrm{base}})
    +
    \sum_{i=1}^{N_r}
    w_i
    \exp \left(
    -\frac{\|\vect{x}-\vect{c}_i\|_2^2}{2\sigma_i^2}
    \right)
    \right),
    \label{eq:initial_belief_roi}
\end{equation}
where \(\sigma(\cdot)\) is the logistic function, \(p_{\mathrm{base}}\) is a background risk probability, and \(w_i\) and \(\sigma_i\) encode the strength and spatial uncertainty of ROI \(r_i\), respectively. This construction allows confident reports to induce concentrated high-risk regions, while uncertain reports produce broader and weaker belief patterns. Other prior models can also be used as long as they define an initial belief field over \(\Omega\).

The UAV starts from a depot location \(\vect{x}_0 \in \Omega\) and has a finite flight distance budget \(D_m\), representing battery or mission endurance. The mission consists of a high-level route and a set of low-level trajectories. The route determines the order in which the UAV inspects the reported ROIs, possibly with additional auxiliary waypoints introduced to improve coverage. The trajectory between consecutive route waypoints must be dynamically feasible and must respect the allocated path budget. Let the route induce a sequence of edges \(\mathcal{E} = \{ e_1, e_2, \dots, e_K \}\), where each edge \(e_k = (\vect{a}_k,\vect{b}_k)\) connects two consecutive route waypoints. We represent the UAV endurance constraint as a maximum allowable flight distance \(D_m\), which may be derived from battery capacity, expected energy consumption, or mission-time limits. For each edge, this flight distance budget is divided into a nominal travel component \(\bar{L}_k\), which is the distance required to move between the edge endpoints, and a marginal exploration budget \(B_k\), which is the additional distance available for sensing-oriented deviations from the direct connection. The resulting edge-wise path budget is given by \(L_k \leq \bar{L}_k + B_k\) for \(k=1,\dots,K\), with the total mission budget satisfying
\begin{equation}
    \sum_{k=1}^{K} L_k \leq D_m.
    \label{eq:total_budget_constraint}
\end{equation}
This defines the path length budget available for each low-level trajectory while preserving the total mission endurance constraint.

\begin{figure}[htbp]
    \begin{center}
        \includegraphics[width=0.65\linewidth]{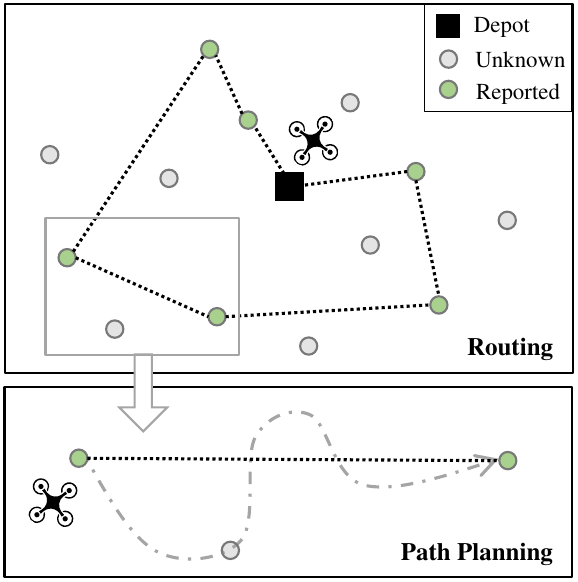}
    \end{center}
    \vspace{-0.5cm}
    \caption{Mission-level concept of operations for ROI-based UAV monitoring. The UAV starts from a depot, visits reported ROIs to satisfy inspection requirements, and uses remaining flight distance budget to explore areas where unreported or mislocalized hazards may exist.}
    \label{fig:conops}
\end{figure}

\Cref{fig:conops} shows the mission-level concept of operations. The depot represents the UAV's base for launch, recovery, charging, data transfer, or coordination. Reported ROIs define inspection requirements that must be included in the high-level route, but they are not treated as confirmed hazard locations. Unknown or unreported hazard regions motivate exploration between reported ROIs. This motivates a bi-level framework in which route optimization determines the sequence of inspection waypoints, while path planning converts each route segment into a dynamically feasible and informative trajectory. The remaining travel capacity after nominal route construction is interpreted as marginal exploration budget and is allocated to route segments for low-level informative planning.

\begin{figure}[htbp]
    \centering
    \includegraphics[width=0.9\linewidth]{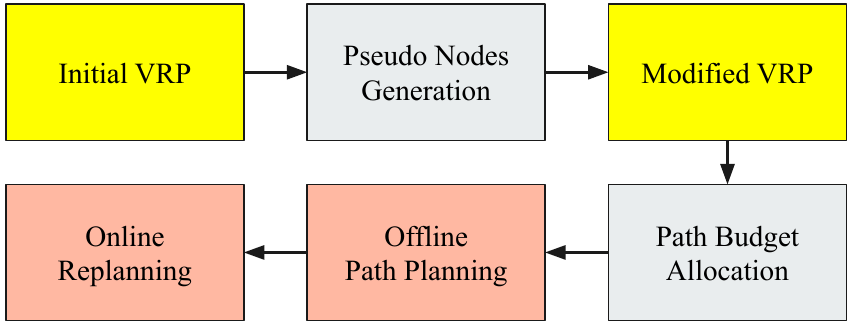}
    \caption{Algorithmic workflow of the proposed offline routing and online belief-driven replanning framework under imperfect and evolving risk information.}
    \label{fig:online_bilevel_flow}
\end{figure}

While \Cref{fig:conops} illustrates the mission geometry, \Cref{fig:online_bilevel_flow} summarizes the algorithmic workflow. The offline layer first constructs an ROI-constrained route, augments the route with pseudo-nodes to improve spatial coverage, assigns edge-wise exploration budgets, and generates initial risk-aware B-spline trajectories. During execution, the UAV collects measurements, updates its grid-based belief field, evaluates replanning conditions, and replans the remaining trajectory when belief changes and remaining budget justify adaptation. In this way, the framework separates mission-level inspection requirements from trajectory-level information gathering while allowing newly acquired measurements to influence the remaining plan.

During execution, the UAV collects measurements along its trajectory, allowing the mission-relevant risk information available to the planner to evolve over time. At time \(t\), when the UAV is located at \(\vect{x}_t\), it receives a sensor measurement \(y_t\) modeled as a noisy local observation of the underlying risk field:
\begin{equation}
    y_t =
    \mu_0
    +
    \frac{\kappa
    \int_{\Omega_t}
    K(\|\vect{x}_t-\vect{\xi}\|) p_{\mathrm{true}}(\vect{\xi}) d\vect{\xi}
    }{
    \int_{\Omega_t}
    K(\|\vect{x}_t-\vect{\xi}\|) d\vect{\xi}
    }
    +
    \epsilon_t,
    \label{eq:continuous_measurement_model}
\end{equation}
where \(\mu_0\) is a baseline signal level, \(\kappa\) is the signal strength, \(K(\cdot)\) is a distance-dependent sensing kernel of the form \(K(d) = \exp(-\beta_s d^2)\), \(\Omega_t \subseteq \Omega\) is the local sensing region, and \(\varepsilon_t\) represents measurement noise. In this work, the continuous domain \(\Omega\) is discretized into a finite set of grid points, and the risk belief is maintained at those locations. This grid-based representation approximates the local sensing integrals and supports efficient belief updates during execution. Other spatial representations, such as Gaussian processes, radial basis functions, particle maps, or adaptive meshes, could be used in place of the grid. The implementation details of the measurement and belief update model are given in \Cref{sec:sensor_belief_model}. The resulting belief field provides the information state used by the downstream planning and replanning layers.

The goal of the mission is to inspect all reported ROIs while improving the UAV's estimate of the continuous risk field. In evaluation, we measure this improvement using the reduction in Kullback-Leibler (KL) divergence between the true risk field and the UAV belief:
\begin{equation}
    \Delta_{\mathrm{KL}}
    =
    D_{\mathrm{KL}}(p_{\mathrm{true}} \,\|\, b_0)
    -
    D_{\mathrm{KL}}(p_{\mathrm{true}} \,\|\, b_T),
    \label{eq:kl_reduction}
\end{equation}
where \(D_{\mathrm{KL}}(\cdot \,\|\, \cdot)\) denotes KL divergence and \(b_T\) is the final belief after the mission. Since \(p_{\mathrm{true}}\) is unavailable to the UAV, \(\Delta_{\mathrm{KL}}\) is used only for evaluation and not as a planning objective. During execution, the planner instead relies on belief-based quantities available from the current belief and uncertainty estimates to generate and evaluate informative trajectories.

The overall problem is therefore to design an integrated monitoring framework that: (i) constructs an ROI-constrained route that enforces inspection of reported regions, (ii) allocates the available flight distance budget across route segments to support exploration, (iii) generates dynamically feasible informative trajectories over the continuous domain, and (iv) updates and replans those trajectories online as new UAV measurements refine the belief field and reveal discrepancies between the prior reports and the sensed environment. The following sections describe the routing layer, the edge-wise exploration budget allocation, the offline informative path planner, and the online belief update and replanning strategy.

\section{High-Level Routing Strategy}
\label{sec:high_level_routing}

The high-level routing layer determines an inspection route that visits all reported ROIs while preserving flight budget for exploration over the continuous risk field. Unlike target-based routing problems that treat nodes as confirmed hazard locations, the required inspection nodes in this work represent reported ROI centers. To support both ROI inspection and exploration, the routing layer first solves a vehicle routing problem with ROIs, then augments the route with optional pseudo-nodes to improve spatial coverage, and finally assigns the remaining flight distance budget to route segments for edge-wise exploration.

\subsection{ROI-Constrained Route Optimization}
\label{sec:roi_vrp}

Let \(N_r\) be the number of reported ROIs and define \(I_r = \{1,2,\dots,N_r\}\) as the corresponding index set. The index \(0\) represents the depot node, and the set of all required routing nodes is given by \(I_0 = I_r \cup \{0\}\). Each ROI node \(i \in I_r\) is associated with the center location \(\vect{c}_i \in \Omega\), and the depot is located at \(\vect{c}_0 \in \Omega\). The distance between nodes \(i,j \in I_0\) is denoted by \(d_{ij}\). The set of UAVs is denoted by \(M\), where each UAV \(m \in M\) has a maximum flight distance budget \(D_m\).

We use a binary decision variable \(y_{ijm} \in \{0,1\}\) to indicate whether UAV \(m\) travels directly from node \(i\) to node \(j\). To eliminate subtours, we introduce a continuous auxiliary variable \(u_{im} \in \mathbb{R}_+\), representing the relative visit order of ROI node \(i\) by UAV \(m\). The ROI-constrained VRP is formulated as
\allowdisplaybreaks
\begin{subequations}
\label{eq:roi_vrp}
\begin{align}
    \min \quad
    & \sum_{m \in M} \sum_{i \in I_0} \sum_{j \in I_0} d_{ij} y_{ijm}
    \label{eq:roi_vrp_obj} \\
    \text{s.t.} \quad
    & \sum_{m \in M} \sum_{\substack{j \in I_0 \\ j \neq i}} y_{ijm} = 1,
    && \forall i \in I_r,
    \label{eq:roi_vrp_visit} \\
    & \sum_{\substack{j \in I_0 \\ j \neq i}} y_{ijm}
    =
    \sum_{\substack{j \in I_0 \\ j \neq i}} y_{jim},
    && \forall m \in M,\; \forall i \in I_0,
    \label{eq:roi_vrp_flow} \\
    & \sum_{i \in I_0} \sum_{j \in I_0} d_{ij} y_{ijm}
    \leq D_m,
    && \forall m \in M,
    \label{eq:roi_vrp_budget} \\
    & \sum_{j \in I_r} y_{0jm} = 1,
    && \forall m \in M,
    \label{eq:roi_vrp_start} \\
    & \sum_{i \in I_r} y_{i0m} = 1,
    && \forall m \in M,
    \label{eq:roi_vrp_end} \\
    & u_{0m} = 0,
    && \forall m \in M,
    \label{eq:roi_vrp_mtz_base} \\
    & 1 \leq u_{im} \leq |I_r|,
    && \forall i \in I_r,\; \forall m \in M,
    \label{eq:roi_vrp_mtz_bounds} \\
    & u_{jm} \geq u_{im} + 1 - |I_r|(1-y_{ijm}),
    && \forall m \in M,\; \forall i \neq j \in I_r,
    \label{eq:roi_vrp_mtz} \\
    & y_{ijm} \in \{0,1\},
    && \forall i,j \in I_0,\; \forall m \in M,
    \label{eq:roi_vrp_binary} \\
    & u_{im} \in \mathbb{R}_+,
    && \forall i \in I_r,\; \forall m \in M.
    \label{eq:roi_vrp_u_domain}
\end{align}
\end{subequations}

Here, \(D_m\) denotes the total flight distance budget of UAV \(m\). The VRP ensures that the high-level ROI route remains feasible within this budget. The objective in \Cref{eq:roi_vrp_obj} minimizes the total route length. \Cref{eq:roi_vrp_visit} ensures that each reported ROI is inspected exactly once, while \Cref{eq:roi_vrp_flow} enforces route continuity. \Cref{eq:roi_vrp_budget} limits the route length for each UAV. \Cref{eq:roi_vrp_start,eq:roi_vrp_end} require each route to start and end at the depot, and \Cref{eq:roi_vrp_mtz_base,eq:roi_vrp_mtz_bounds,eq:roi_vrp_mtz} eliminate subtours using the Miller--Tucker--Zemlin formulation~\cite{mtz}.

\subsection{Pseudo-Node Route Augmentation}
\label{sec:pseudo_node_augmentation}

The route obtained from the ROI-constrained VRP guarantees inspection of the reported ROIs, but it may provide poor spatial coverage when the ROIs are clustered or unevenly distributed. Since the environment is represented as a continuous risk field, the path between ROIs can provide important sensing opportunities. To improve the spatial distribution of the route, we introduce pseudo-nodes as auxiliary coverage waypoints. These nodes are not hazard reports and do not introduce additional ROI inspection requirements. Instead, they are mission-design waypoints that can be added or removed depending on the desired coverage and available flight budget.

Pseudo-nodes should improve coverage by drawing the route toward regions that are poorly represented by the current ROI route. We use a centroidal Voronoi tessellation (CVT)~\cite{cvt} procedure because it provides a simple way to distribute auxiliary points according to a chosen spatial density. By assigning larger density to regions farther from the existing route edges, the CVT update places pseudo-nodes in under-covered areas, as illustrated in \Cref{fig:routing_strategy_sim}. Given a set of generators \(z_i \in \Omega\), the Voronoi cell associated with generator \(z_i\) is
\begin{equation}
    V_i =
    \left\{
    \vect{x} \in \Omega
    \mid
    \|\vect{x}-z_i\|_2
    \leq
    \|\vect{x}-z_j\|_2,\; \forall j \neq i
    \right\}.
    \label{eq:cvt_cell}
\end{equation}
In a standard CVT, each generator is updated to the centroid of its Voronoi cell
\begin{equation}
    z_i =
    \frac{
    \int_{V_i} \vect{x}\rho(\vect{x})d\vect{x}
    }{
    \int_{V_i}\rho(\vect{x})d\vect{x}
    },
    \label{eq:cvt_centroid}
\end{equation}
where \(\rho(\vect{x})\) is a density function over the domain. With a uniform density, the resulting generators tend to be evenly distributed over \(\Omega\).

To account for the current route structure, we use an edge-based density function that assigns higher weight to regions farther from the existing route edges. The intuition is that locations close to the current route are more likely to be sensed by trajectories generated along that route, whereas locations far from the route are more likely to remain under-covered. In the CVT update, each pseudo-node moves toward the weighted centroid of the region assigned to it. Therefore, if an assigned region contains many points far from the current route, its weighted centroid is pulled in that direction. This causes pseudo-nodes to move toward under-covered regions while keeping the reported ROI centers fixed.

Let \(\mathcal{E}_{\mathrm{route}}\) denote the set of edges in the current route. The density is defined as
\begin{equation}
    \rho(\vect{x})
    =
    \alpha_{\rho} + \beta_{\rho} d(\vect{x},\mathcal{E}_{\mathrm{route}}),
    \label{eq:edge_cvt_density}
\end{equation}
where \(d(\vect{x},\mathcal{E}_{\mathrm{route}})\) is the Euclidean distance from \(\vect{x}\) to the closest route edge, and \(\alpha_{\rho},\beta_{\rho}>0\) are weighting parameters. Points farther from the current route therefore contribute more strongly to the weighted centroids in the CVT update. During the update, ROI centers are treated as fixed generators so that the required inspection locations remain unchanged, while only pseudo-nodes are moved. The resulting pseudo-nodes guide the augmented route toward under-covered regions without changing the reported ROI inspection requirements.

After pseudo-node generation, the VRP is re-solved over the augmented node set \(\mathcal{Z}=\{\vect{c}_i\}_{i=1}^{N_r} \cup \mathcal{P}\), where \(\mathcal{P}\) is the set of pseudo-nodes. In the augmented VRP, both ROI centers and selected pseudo-nodes are included as route waypoints. The distinction is that ROI centers represent required inspection regions, whereas pseudo-nodes are mission-design waypoints introduced solely to improve route-level spatial coverage. Thus, pseudo-node visitation is mandatory only after the mission designer selects \(N_p\), and setting \(N_p=0\) recovers the ROI-only route. The resulting route preserves ROI inspection while improving spatial coverage for exploration.

Adding pseudo-nodes introduces a tradeoff between route-level spatial coverage and remaining marginal exploration budget. More pseudo-nodes can guide the route through under-covered regions and create more informative edge geometries, but they also increase the nominal route length and may reduce \(B_{\mathrm{marg}}\). The number of pseudo-nodes \(N_p\) is therefore treated as a mission-design parameter. In the experiments, we vary \(N_p\) to evaluate this tradeoff, including the ROI-only case \(N_p=0\).

\begin{algorithm}[htbp]
\caption{Edge-based CVT Pseudo-Node Generation}
\label{alg:edge_aware_cvt}
\begin{algorithmic}[1]
    \Require ROI centers \(\{\vect{c}_i\}_{i=1}^{N_r}\), current route edges \(\mathcal{E}_{\mathrm{route}}\), domain \(\Omega\), number of pseudo-nodes \(N_p\), weights \(\alpha_{\rho},\beta_{\rho}\)
    \State Initialize pseudo-nodes \(\mathcal{P}=\{\vect{p}_1,\dots,\vect{p}_{N_p}\}\) in \(\Omega\)
    \For{\(h=1\) to \(h_{\max}\)}
        \State Sample points \(\mathcal{S}\subset\Omega\)
        \State Assign each sampled point to the nearest generator in \(\{\vect{c}_i\}_{i=1}^{N_r}\cup \mathcal{P}\)
        \For{each pseudo-node \(\vect{p}_\ell \in \mathcal{P}\)}
            \State Update \(\vect{p}_\ell\) using the weighted centroid in \Cref{eq:cvt_centroid} with density \Cref{eq:edge_cvt_density}
        \EndFor
    \EndFor
    \State Re-solve the VRP over the augmented node set \(\mathcal{Z}\)
    \State \Return augmented route and route edges
\end{algorithmic}
\end{algorithm}

\subsection{Edge-Wise Exploration Budget Allocation}
\label{sec:edge_budget_allocation}

After the augmented route is computed, the remaining flight distance budget must be assigned to individual route segments so that the path planner can use it for local exploration. Rather than giving the low-level planner a single global budget, we define an edge-wise budget for each route edge. This specifies the maximum trajectory length available on that edge and ensures that the local trajectory plans remain consistent with the mission-level flight distance budget.

Let \(\mathcal{E}=\{e_1,\dots,e_K\}\) be the set of route edges after pseudo-node augmentation. Each edge \(e_k=(\vect{a}_k,\vect{b}_k)\) connects two consecutive waypoints in the route. The nominal travel length \(\bar{L}_k\) represents the minimum feasible travel required to move from \(\vect{a}_k\) to \(\vect{b}_k\). In the simplest case, \(\bar{L}_k\) can be the Euclidean distance between the endpoints. For curvature-constrained vehicles, it can be computed using a kinematically feasible surrogate such as a Dubins shortest path length~\cite{dubins}.

The total marginal budget available for exploration is
\begin{equation}
    B_{\mathrm{marg}}
    =
    D_m
    -
    \sum_{k=1}^{K} \bar{L}_k.
    \label{eq:marginal_budget}
\end{equation}
Thus, exploration does not use an additional budget outside the UAV endurance limit. Instead, it uses only the remaining distance budget after the augmented route waypoints have been connected by nominal feasible travel segments. Assuming \(B_{\mathrm{marg}}\geq 0\), this remaining budget is distributed across route edges using a line-segment Voronoi partition of the domain. This partition assigns each location in \(\Omega\) to the route edge that is closest to it. Intuitively, an edge whose surrounding region covers a larger portion of the domain receives more exploration allowance, since trajectories along that edge are responsible for sensing a larger local area. The Voronoi region associated with edge \(e_k\) is
\begin{equation}
    V_k =
    \left\{
    \vect{x}\in\Omega
    \mid
    d(\vect{x},e_k)
    \leq
    d(\vect{x},e_l),\; \forall l\neq k
    \right\}.
    \label{eq:edge_voronoi_cell}
\end{equation}
Here, \(d(\vect{x},e_k)\) is the shortest distance from \(\vect{x}\) to the finite line segment connecting \(\vect{a}_k\) and \(\vect{b}_k\), rather than to the infinite line through the two endpoints. The scalar \(\eta\) gives the normalized position of the orthogonal projection of \(\vect{x}\) along the segment: \(\eta=0\) corresponds to \(\vect{a}_k\), \(\eta=1\) corresponds to \(\vect{b}_k\), and \(0 \leq \eta \leq 1\) means the projection lies between the endpoints. If \(\eta<0\) or \(\eta>1\), the projection falls outside the segment, so the nearest point is the corresponding endpoint. For \(e_k=(\vect{a}_k,\vect{b}_k)\), the projection factor is
\begin{equation}
    \eta
    =
    \frac{
    (\vect{x}-\vect{a}_k)^\top(\vect{b}_k-\vect{a}_k)
    }{
    \|\vect{b}_k-\vect{a}_k\|_2^2
    },
    \label{eq:projection_factor}
\end{equation}
and
\begin{equation}
d(\vect{x},e_k)=
\begin{cases}
\|\vect{x}-\vect{a}_k\|_2, & \eta < 0, \\
\|\vect{x}-\vect{b}_k\|_2, & \eta > 1, \\
\|\vect{x}-(\vect{a}_k+\eta(\vect{b}_k-\vect{a}_k))\|_2, & \text{otherwise}.
\end{cases}
\label{eq:point_segment_distance}
\end{equation}

Let \(A_k\) be the area of \(V_k\). A simple area-based allocation assigns
\begin{equation}
    B_k
    =
    B_{\mathrm{marg}}
    \frac{A_k}{\sum_{\ell=1}^{K} A_\ell}.
    \label{eq:area_budget_allocation}
\end{equation}
More generally, the allocation may include a belief-dependent score \(s_k\) computed over \(V_k\). The score \(s_k\) may be chosen as the average current belief, average uncertainty, or a weighted combination of risk and uncertainty over \(V_k\). The resulting edge weight is \(\omega_k = A_k s_k\), and the budget assignment is
\begin{equation}
    B_k
    =
    \frac{B_{\mathrm{marg}}\omega_k}{\sum_{\ell=1}^{K} \omega_\ell}.
    \label{eq:belief_weighted_budget_allocation}
\end{equation}
The resulting flight distance budget for edge \(e_k\) is
\begin{equation}
    L_k \leq \bar{L}_k + B_k.
    \label{eq:routing_length_budget}
\end{equation}
This budget is passed to the low-level path planner, which generates a dynamically feasible informative trajectory between \(\vect{a}_k\) and \(\vect{b}_k\) within the assigned budget.

\section{Low-Level Informative Path Planning and Online Replanning}
\label{sec:low_level_planning}

The high-level routing layer provides a sequence of route edges and an edge-wise budget for each segment. The low-level planner converts each route edge into a dynamically feasible trajectory that is informative with respect to the current belief field. Unlike the previous discrete hazard formulation, the planner operates over a continuous risk belief \(b_t(\vect{x})\), which is updated during execution using streaming UAV measurements. This section presents the belief and sensing model, the B-spline trajectory representation, the offline edge-wise path planning problem, the replanning evaluation score, and the online replanning strategy.

\subsection{Sensor Model and Belief Representation}
\label{sec:sensor_belief_model}

The continuous domain \(\Omega\) is represented by a finite set of grid points \(\mathcal{G} = \{\vect{g}_1,\vect{g}_2,\dots,\vect{g}_{N_g}\}\), where each \(\vect{g}_i \in \Omega\) is a grid location at which the risk belief is evaluated. At time \(t\), the UAV maintains a belief probability \(b_t(\vect{g}_i) \in [0,1]\) and represents the same belief using the log-odds representation
\begin{equation}
    \ell_t(\vect{g}_i)
    =
    \operatorname{logit}(b_t(\vect{g}_i))
    =
    \log\left(
    \frac{b_t(\vect{g}_i)}{1-b_t(\vect{g}_i)}
    \right).
    \label{eq:log_odds_representation}
\end{equation}
The log-odds representation is commonly used in occupancy-grid mapping~\cite{thrun2002probabilistic} because it converts a bounded probability into an unbounded real-valued quantity, making sequential measurement updates easier to express. The belief can be recovered from the log-odds representation as \(b_t(\vect{g}_i)=\sigma(\ell_t(\vect{g}_i))\), where \(\sigma(\cdot)\) is the logistic function. The planner also maintains a logit-space variance \(v_t(\vect{g}_i)\), which captures the uncertainty associated with the current belief estimate.

When the UAV is located at \(\vect{x}_t\), it receives a scalar measurement \(y_t\). The measurement is modeled as a noisy local observation of the underlying risk field. For belief prediction and update, we define the local update neighborhood
\begin{equation}
    \mathcal{N}_t
    =
    \left\{
    i \in \{1,\dots,N_g\}
    \mid
    \|\vect{x}_t-\vect{g}_i\|_2 \leq r_{\mathrm{upd}}
    \right\},
    \label{eq:update_neighborhood}
\end{equation}
and the sensing kernel
\begin{equation}
    K_{ti}
    =
    \exp\left(-\beta_s \|\vect{x}_t-\vect{g}_i\|_2^2\right),
    \qquad i \in \mathcal{N}_t.
    \label{eq:sensing_kernel}
\end{equation}
Here, \(K_{ti}\) is the sensing weight from the executed UAV position \(\vect{x}_t\) to grid point \(\vect{g}_i\), and \(r_{\mathrm{upd}}\) is the local update radius. Given the current belief, the predicted measurement is
\begin{equation}
    \hat{y}_t
    =
    \mu_0
    +
    \frac{\kappa
    \sum_{i\in\mathcal{N}_t} K_{ti} b_t(\vect{g}_i)
    }{
    \sum_{i\in\mathcal{N}_t} K_{ti}
    },
    \label{eq:predicted_measurement}
\end{equation}
where \(\mu_0\) is the baseline signal level, \(\kappa\) is the signal strength, and \(\beta_s\) controls the spatial decay of the sensor response.

The belief is updated using an extended Kalman filter (EKF) in log-odds space. Let
\begin{equation}
    H_{ti}
    =
    \frac{
    \kappa K_{ti} b_t(\vect{g}_i)(1-b_t(\vect{g}_i))
    }{
    \sum_{j\in\mathcal{N}_t}K_{tj}
    },
    \qquad i \in \mathcal{N}_t,
    \label{eq:ekf_jacobian}
\end{equation}
be the linearized observation Jacobian with respect to the log-odds representation. The innovation variance is
\begin{equation}
    S_t
    =
    \sigma_y^2
    +
    \sum_{i\in\mathcal{N}_t}
    v_t(\vect{g}_i)H_{ti}^2,
    \label{eq:innovation_variance}
\end{equation}
where \(\sigma_y\) is the measurement noise standard deviation. The Kalman gain for grid point \(\vect{g}_i\) is
\begin{equation}
    G_{ti}
    =
    \frac{
    v_t(\vect{g}_i)H_{ti}
    }{
    S_t
    },
    \qquad i \in \mathcal{N}_t.
    \label{eq:kalman_gain}
\end{equation}
The log-odds representation and variance are updated as
\begin{align}
    \ell_{t+1}(\vect{g}_i)
    &=
    \ell_t(\vect{g}_i)
    +
    G_{ti}(y_t-\hat{y}_t),
    \qquad i \in \mathcal{N}_t,
    \label{eq:log_odds_update} \\
    v_{t+1}(\vect{g}_i)
    &=
    v_t(\vect{g}_i)
    -
    G_{ti}H_{ti}v_t(\vect{g}_i),
    \qquad i \in \mathcal{N}_t.
    \label{eq:variance_update}
\end{align}
For grid points outside \(\mathcal{N}_t\), the belief and variance remain unchanged except for optional spatial diffusion. A lightweight spatial diffusion step may be applied after the local update to propagate measurement influence to nearby cells and reduce over-confident localization caused by sparse measurements.

\subsection{B-Spline Trajectory Representation}
\label{sec:bspline_trajectory}

For each route edge \(e_k=(\vect{a}_k,\vect{b}_k)\), the low-level planner generates a continuous trajectory from the start waypoint \(\vect{a}_k\) to the end waypoint \(\vect{b}_k\). We parameterize the trajectory using a B-spline since it provides a smooth and differentiable representation suitable for constrained trajectory optimization.

Let \(\mathcal{C}_k = \{\vect{c}_{k,1},\vect{c}_{k,2},\dots,\vect{c}_{k,N_c}\}\) denote the control points for edge \(e_k\), and let \(r\) denote the B-spline order. The trajectory is written as
\begin{equation}
    \vect{p}_k(\tau)
    =
    \sum_{i=1}^{N_c}
    B_{i,r}(\tau)\vect{c}_{k,i},
    \qquad \tau \in [0,t_{f,k}],
    \label{eq:bspline_path}
\end{equation}
where \(B_{i,r}(\tau)\) are B-spline basis functions, \(\tau\) is the trajectory parameter, and \(t_{f,k}\) is the final traversal time for edge \(e_k\). The derivatives of \(\vect{p}_k(\tau)\) are used to impose vehicle motion constraints.

We use a planar unicycle model as a reduced-order representation of UAV motion. This model captures the horizontal motion constraints most relevant to the proposed planning layer, including forward speed and heading-rate limits, while avoiding the computational cost of full quadrotor dynamics. More detailed models could be used when attitude dynamics, thrust limits, or three-dimensional maneuvering are critical. Here, the unicycle abstraction is appropriate because the planner operates at the path-planning level, assumes approximately constant-altitude monitoring, and relies on a lower-level flight controller to track the planned trajectory. Accordingly, the UAV is modeled using unicycle kinematics:
\begin{equation}
\begin{bmatrix}
\dot{x}(\tau) \\
\dot{y}(\tau) \\
\dot{\theta}(\tau)
\end{bmatrix}
=
\begin{bmatrix}
v(\tau)\cos\theta(\tau) \\
v(\tau)\sin\theta(\tau) \\
u(\tau)
\end{bmatrix},
\label{eq:unicycle_model_low_level}
\end{equation}
where \(x(\tau)\) and \(y(\tau)\) are the planar position coordinates, \(\theta(\tau)\) is the heading angle, \(v(\tau)\) is the speed, and \(u(\tau)\) is the turn rate. For a B-spline trajectory, the corresponding speed, turn rate, and curvature are obtained from the first and second derivatives of \(\vect{p}_k(\tau)\):
\begin{equation}
\begin{aligned}
    v(\tau)
    &= \|\dot{\vect{p}}_k(\tau)\|_2,
    \quad
    u(\tau)
    =
    \frac{
    \dot{\vect{p}}_k(\tau) \times \ddot{\vect{p}}_k(\tau)
    }{
    \|\dot{\vect{p}}_k(\tau)\|_2^2
    },
    \quad
    \chi(\tau)
    =
    \frac{u(\tau)}{v(\tau)} .
\end{aligned}
\label{eq:spline_kinematic_quantities}
\end{equation}
Here, \(\chi(\tau)\) denotes curvature, and \(\times\) denotes the scalar cross product in two dimensions. The trajectory length is
\begin{equation}
    L(\vect{p}_k)
    =
    \int_0^{t_{f,k}}
    \|\dot{\vect{p}}_k(\tau)\|_2 d\tau.
    \label{eq:spline_length}
\end{equation}

\subsection{Offline Edge-Wise Informative Path Planning}
\label{sec:offline_informative_planning}

The offline low-level planner generates an initial trajectory for each route edge using the belief field available before executing that edge. For notational consistency, this belief is denoted by \(b_t\), where \(t\) represents the planning time for the current edge. Let \(\mathcal{G}_k \subseteq \mathcal{G}\) denote the set of grid points assigned to edge \(e_k\), for example through the edge-wise Voronoi partition described in \Cref{sec:edge_budget_allocation}. The path planner receives the edge endpoints \((\vect{a}_k,\vect{b}_k)\), the current belief \(b_t\), and the edge-wise budget \(L(\vect{p}_k) \leq \bar{L}_k+B_k\).

To construct an informative trajectory, we use a belief-weighted expected detection objective over the grid points assigned to the edge. This objective is a differentiable planning surrogate that encourages sensing near high-belief regions while remaining compatible with constrained B-spline optimization.

Let \(\mathcal{T}_k=\{\tau_1,\tau_2,\dots,\tau_{N_s}\}\) be the set of sampling times along the candidate trajectory, and define \(\vect{s}_{k,j}=\vect{p}_k(\tau_j)\). Here, \(j\) indexes trajectory samples, while \(i\) indexes grid points. For grid point \(\vect{g}_i\), let \(d_{ij}=\|\vect{s}_{k,j}-\vect{g}_i\|_2\) and define the smooth sensing weight \(K_{ij}=\exp(-\beta_s d_{ij}^2)\). The same exponential kernel form is used for belief updates and trajectory evaluation: \(K_{ti}\) is evaluated at the executed UAV position \(\vect{x}_t\), whereas \(K_{ij}\) is evaluated at candidate trajectory sample \(\vect{s}_{k,j}\). The expected signal level induced by risk at \(\vect{g}_i\) is
\begin{equation}
    \mu_1(d_{ij})
    =
    \mu_0
    +
    \kappa K_{ij}.
    \label{eq:expected_signal_level}
\end{equation}
Although the belief update uses the continuous scalar measurement \(y_t\), the path objective evaluates whether a candidate trajectory is likely to produce informative sensor responses by thresholding the predicted signal distribution. Using the detection threshold \(y_{\mathrm{th}}\), the probability of receiving an informative detection response associated with grid point \(\vect{g}_i\) at sample location \(\vect{s}_{k,j}\) is modeled as
\begin{equation}
    \pi_{ij}
    =
    1
    -
    \Phi_{\mathcal{N}}
    \left(
    \frac{y_{\mathrm{th}}-\mu_1(d_{ij})}{\sigma_y}
    \right),
    \label{eq:single_sample_detection_prob}
\end{equation}
where \(\Phi_{\mathcal{N}}(\cdot)\) is the standard normal cumulative distribution function. Since \(\mu_1(d_{ij})\) is defined through the smooth sensing weight \(K_{ij}\), the resulting detection probability remains differentiable with respect to the trajectory sample locations, preserving differentiability for IPOPT-based constrained optimization.

Using a standard independent response approximation across samples, the cumulative informative-response score associated with \(\vect{g}_i\) along the path is
\begin{equation}
    S_{\mathrm{det}}(\vect{g}_i \mid \vect{p}_k)
    =
    1
    -
    \prod_{j=1}^{N_s}
    (1-\pi_{ij}).
    \label{eq:path_detection_score}
\end{equation}
The expected detection score of the trajectory is then
\begin{equation}
    \Psi_k(\vect{p}_k;b_t)
    =
    \sum_{\vect{g}_i \in \mathcal{G}_k}
    b_t(\vect{g}_i)
    S_{\mathrm{det}}(\vect{g}_i \mid \vect{p}_k).
    \label{eq:expected_detection_objective}
\end{equation}
This score is not a posterior divergence objective. Direct optimization of the evaluation metric \(\Delta_{\mathrm{KL}}\) defined in \Cref{eq:kl_reduction} would require access to \(p_{\mathrm{true}}\), which is unavailable during execution, while exact posterior information-gain optimization would substantially increase the complexity of the constrained B-spline problem. The proposed score therefore serves as a differentiable surrogate that biases the trajectory toward high-belief regions with strong predicted sensor response.

The endpoint velocities \(\vect{v}_{k,0}\) and \(\vect{v}_{k,f}\) define the desired entry and exit velocities for edge \(e_k\). The constants \(v_{\min}, v_{\max}, u_{\min}, u_{\max}\), and \(\chi_{\max}\) define the speed, turn-rate, and curvature bounds. The offline path planning problem for edge \(e_k\) is formulated as
\begin{subequations}
\label{eq:offline_path_optimization}
\begin{align}
    \mathcal{C}_k^\star,t_{f,k}^\star
    &=
    \operatorname*{argmin}_{\mathcal{C}_k,t_{f,k}}
    -\Psi_k(\vect{p}_k;b_t)
    \label{eq:offline_path_obj} \\
    \text{s.t.} \quad
    & \vect{p}_k(0) = \vect{a}_k,\quad
      \vect{p}_k(t_{f,k}) = \vect{b}_k,
    \label{eq:offline_pos_constraints} \\
    & \dot{\vect{p}}_k(0) = \vect{v}_{k,0},\quad
      \dot{\vect{p}}_k(t_{f,k}) = \vect{v}_{k,f},
    \label{eq:offline_vel_constraints} \\
    & v_{\min} \leq v(\tau_j) \leq v_{\max},
    \quad \forall \tau_j \in \mathcal{T}_k,
    \label{eq:offline_speed_constraints} \\
    & u_{\min} \leq u(\tau_j) \leq u_{\max},
    \quad \forall \tau_j \in \mathcal{T}_k,
    \label{eq:offline_turn_constraints} \\
    & -\chi_{\max} \leq \chi(\tau_j) \leq \chi_{\max},
    \quad \forall \tau_j \in \mathcal{T}_k,
    \label{eq:offline_curvature_constraints} \\
    & L(\vect{p}_k) \leq \bar{L}_k + B_k.
    \label{eq:offline_length_constraint}
\end{align}
\end{subequations}
The first two constraints enforce the edge endpoints and endpoint velocities. The next three constraints impose speed, turn-rate, and curvature limits at the sampled trajectory points. The final constraint enforces the edge-wise budget from the high-level routing layer.

\subsection{Information-Gain Score for Replanning}
\label{sec:expected_ig_score}

Before describing the online replanning trigger and acceptance rule, we define the expected information-gain score used only to evaluate candidate replans. The expected detection score in \Cref{eq:expected_detection_objective} is used as the primary objective for offline trajectory optimization. During online execution, however, replanning also requires deciding whether a candidate trajectory is sufficiently better than the remaining portion of the current trajectory. For this acceptance test, we use a separate risk-weighted expected information-gain score based on the current belief and logit-space variance.

For a candidate trajectory \(\vect{p}_k\), let \(K_{ij}\) denote the sensing kernel between sample \(\vect{p}_k(\tau_j)\) and grid point \(\vect{g}_i\), and let \(\mathcal{N}_{j}\) denote the set of grid points within the local update neighborhood of \(\vect{p}_k(\tau_j)\). For \(\ell \in \mathcal{N}_j\), \(K_{\ell j}=\exp(-\beta_s\|\vect{p}_k(\tau_j)-\vect{g}_\ell\|_2^2)\). Here, \(j\) indexes trajectory samples, while \(i\) and \(\ell\) index grid points. The normalized linearized measurement sensitivity is defined as
\begin{equation}
    H_{ij}
    =
    \frac{
    \kappa K_{ij} b_t(\vect{g}_i)(1-b_t(\vect{g}_i))
    }{
    \sum_{\ell \in \mathcal{N}_{j}} K_{\ell j}
    },
    \qquad i \in \mathcal{N}_{j}.
    \label{eq:ig_jacobian}
\end{equation}
For grid points outside \(\mathcal{N}_j\), we set \(H_{ij}=0\). The accumulated information for grid point \(\vect{g}_i\) is
\begin{equation}
    \mathcal{I}_i(\vect{p}_k)
    =
    \frac{1}{\sigma_y^2}
    \sum_{\tau_j\in\mathcal{T}_k:\, i\in \mathcal{N}_j}
    H_{ij}^2.
    \label{eq:accumulated_information}
\end{equation}
The expected information-gain score is
\begin{equation}
    \mathcal{J}_k(\vect{p}_k;b_t,v_t)
    =
    \sum_{\vect{g}_i\in\mathcal{G}_k}
    b_t(\vect{g}_i)
    \frac{1}{2}
    \log
    \left(
    1
    +
    v_t(\vect{g}_i)
    \mathcal{I}_i(\vect{p}_k)
    \right).
    \label{eq:expected_information_gain}
\end{equation}
This score gives greater value to trajectories that sense regions with both high-risk belief and high remaining uncertainty. It is not used as the primary offline B-spline optimization objective; instead, it evaluates whether an online candidate replan provides additional expected value relative to the current remaining trajectory.

\subsection{Online Belief Update and Replanning}
\label{sec:online_replanning}

The offline trajectory is executed while the UAV continuously collects measurements and updates its belief. After each measurement, the belief update in \Cref{eq:log_odds_update,eq:variance_update} is applied to the local neighborhood of the UAV. The updated belief is then used to determine whether the remaining trajectory should be replanned.

Let \(b_{\mathrm{ref}}\) denote the belief snapshot used as the reference for the current trajectory. The cumulative belief change over the edge grid is measured as
\begin{equation}
    \Delta b_t
    =
    \frac{1}{|\mathcal{G}_k|}
    \sum_{\vect{g}_i\in\mathcal{G}_k}
    \left|
    b_t(\vect{g}_i)
    -
    b_{\mathrm{ref}}(\vect{g}_i)
    \right|.
    \label{eq:belief_change_metric}
\end{equation}
A replanning opportunity is considered only when two conditions hold. First, the belief change must satisfy \(\Delta b_t \geq \varepsilon_b\), where \(\varepsilon_b\) is the belief-change threshold. This prevents replanning from being triggered by small measurement fluctuations. Second, sufficient flight distance budget must remain to reach the edge endpoint while still allowing useful exploration. Let \(D_{k,\mathrm{rem}}\) be the remaining flight distance budget for edge \(e_k\), and let \(\bar{L}_{k,\mathrm{rem}}\) denote the nominal feasible travel length from the current UAV position to the edge endpoint. The remaining budget \(D_{k,\mathrm{rem}}\) is updated by subtracting the distance already flown on the current edge from the edge-wise budget \(\bar{L}_k+B_k\). Replanning is allowed only if
\begin{equation}
    D_{k,\mathrm{rem}}
    \geq
    \bar{L}_{k,\mathrm{rem}}
    +
    \delta,
    \label{eq:remaining_budget_replanning}
\end{equation}
where \(\delta > 0\) is the minimum exploration margin required to justify replanning. Together, these conditions ensure that replanning occurs only when the belief has changed meaningfully and enough path budget remains to act on the new information.

When these conditions are satisfied, a new B-spline trajectory is generated from the current UAV position to the edge endpoint using the updated belief field. The replanning problem has the same structure as \Cref{eq:offline_path_optimization}, but its start state, remaining length budget, and belief field are replaced by their current execution-time values. The new trajectory is accepted only if its expected information gain improves over the remaining portion of the current trajectory:
\begin{equation}
    \mathcal{J}_k(\vect{p}_{\mathrm{new}};b_t,v_t)
    \geq
    \mathcal{J}_k(\vect{p}_{\mathrm{old,rem}};b_t,v_t)
    +
    \delta_{\psi},
    \label{eq:replan_acceptance}
\end{equation}
where \(\delta_{\psi}\geq 0\) is an improvement threshold. If the candidate is accepted, the remaining trajectory is replaced by the new B-spline; if it is rejected, the UAV continues executing the existing trajectory. After each replanning evaluation, regardless of acceptance, the reference belief \(b_{\mathrm{ref}}\) is reset to the current belief to avoid repeated triggers from the same measurement-induced belief change. The acceptance criterion is therefore deliberately separated from the offline path objective. The offline optimizer uses the belief-weighted expected detection score to generate feasible informative trajectories, while the online acceptance test uses the risk-weighted expected information-gain score to decide whether replacing the remaining trajectory is justified under the updated belief and variance.

\begin{algorithm}[htbp]
\caption{Online Belief Update and Replanning for Edge}
\label{alg:online_replanning}
\begin{algorithmic}[1]
    \Require Edge \(e_k=(\vect{a}_k,\vect{b}_k)\), grid \(\mathcal{G}_k\), belief \(b_t\), variance \(v_t\), flight distance budget \(\bar{L}_k+B_k\)
    \State Generate initial B-spline trajectory by solving \Cref{eq:offline_path_optimization}
    \State Set \(b_{\mathrm{ref}} \leftarrow b_t\)
    \While{the UAV has not reached \(\vect{b}_k\)}
        \State Execute the current trajectory for one sampling step
        \State Collect measurement \(y_t\)
        \State Update \(\ell_t\), \(b_t\), and \(v_t\) using \Cref{eq:log_odds_update,eq:variance_update}
        \State Compute belief change \(\Delta b_t\) using \Cref{eq:belief_change_metric}
        \If{\(\Delta b_t \geq \varepsilon_b\) and \(D_{k,\mathrm{rem}} \geq \bar{L}_{k,\mathrm{rem}}+\delta\)}
            \State Replan a B-spline from the current state to \(\vect{b}_k\)
            \If{\(\mathcal{J}_k(\vect{p}_{\mathrm{new}};b_t,v_t) \geq \mathcal{J}_k(\vect{p}_{\mathrm{old,rem}};b_t,v_t)+\delta_{\psi}\)}
                \State Replace the remaining trajectory with \(\vect{p}_{\mathrm{new}}\)
            \Else
                \State Continue executing the existing trajectory
            \EndIf
            \State Reset \(b_{\mathrm{ref}} \leftarrow b_t\)
        \EndIf
    \EndWhile
    \State \Return updated belief field and executed trajectory
\end{algorithmic}
\end{algorithm}

The procedure in Algorithm~\ref{alg:online_replanning} is applied sequentially to each route edge. After completing edge \(e_k\), the final belief becomes the initial belief for planning and executing the next edge \(e_{k+1}\). The offline planner is recovered as a special case of this procedure by disabling the replanning trigger after the initial edge-wise B-spline trajectories are generated. This provides a natural baseline for evaluating the benefit of online replanning.

\section{Numerical Experiments}
\label{sec:numerical_experiments}

This section evaluates the proposed framework through numerical simulations. The experiments are organized into three parts. \Cref{sec:exp_route_augmentation} evaluates the effect of pseudo-node augmentation on route-level spatial coverage. \Cref{sec:online_replanning_analysis} analyzes the behavior and computational cost of the online replanning layer under a representative scenario. \Cref{sec:path_planning_performance} compares all four path planning strategies across 48 scenario configurations varying the number of total clusters~\(n_c\), hidden clusters~\(n_h\), and pseudo-nodes~\(N_p\).

\subsection{Experimental Setup}
\label{sec:experimental_setup}

All simulations are conducted in a bounded two-dimensional domain $\Omega = [0,1000]\times[0,1000]\; \mathrm{m}^2,$ with the depot node located at $(0,0)$ m. The domain is discretized into a uniform grid for belief representation, sensor updates, and metric evaluation. The true risk field is generated synthetically as a mixture of Gaussian risk clusters. For each scenario, \(n_h\) clusters are hidden from the initial prior and are not reported to the planner. The remaining clusters define reported ROIs, whose center locations are perturbed by Gaussian noise before constructing the initial belief field. This produces imperfect prior information that is spatially imprecise and incomplete. The true risk field is used only for measurement generation and post-mission evaluation, and is not available to the planner.

The combination of \(n_c\), \(n_h\), and \(N_p\) gives \(3 \times 4 \times 4 = 48\) scenarios. For each scenario, we run 100 independent random trials. The same scenario instances are used for all compared methods so that performance differences reflect the planning strategy rather than differences in the generated risk fields. The main simulation parameters are summarized in  \Cref{tab:simulation_parameters}. The vehicle routing problem is solved using PyVRP~\cite{pyvrp}. The B-spline trajectory optimization and online replanning problems are solved using IPOPT~\cite{ipopt}. All reported quantitative results are averaged over 100 independent random trials. Simulations were run on a laptop computer with a 13th Gen Intel(R) Core(TM) i7-1360P CPU.

\begin{table}[htbp]
\caption{Simulation Parameters}
\label{tab:simulation_parameters}
\centering
\begin{tabular}{l c}
\hline
\textbf{Parameter} & \textbf{Value} \\
\hline
Domain size                                    & $1000 \times 1000\;\mathrm{m}^2$ \\
Number of trials                              & 100 \\
UAV flight distance budget $D_m$                       & $5000\;\mathrm{m}$ \\
Number of total risk clusters $n_c$           & $\{6,8,10\}$ \\
Number of hidden clusters $n_h$               & $\{1,2,3,4\}$ \\
Cluster spread $\sigma_{\mathrm{cluster}}$    & $160\;\mathrm{m}$ \\
Number of reported ROIs $N_r$                 & $n_c-n_h$ \\
Number of pseudo-nodes $N_p$                  & $\{0,1,2,3\}$  \\
Prior location noise $\sigma_{\mathrm{noise}}$ & $100\;\mathrm{m}$ \\
Sensor signal strength $\kappa$               & 0.8 \\
Sensor kernel decay $\beta_s$                 & 0.005 \\
Measurement noise std $\sigma_y$              & 0.10 \\
Kernel truncation radius $r_{\mathrm{upd}}$              & $20\;\mathrm{m}$ \\
Belief-change threshold $\varepsilon_b$       & 0.01 \\
Speed range $[v_{\min}, v_{\max}]$            & $[0.5,\;10.0]\;\mathrm{m/s}$ \\
Turn-rate range $[u_{\min}, u_{\max}]$        & $[-2.0,\;2.0]\;\mathrm{rad/s}$ \\
CVT density weight $\alpha_\rho$      & 0.1 \\
CVT density weight $\beta_\rho$       & 1.0 \\
Detection threshold $y_{\mathrm{th}}$ & 0.6 \\
\hline
\end{tabular}
\end{table}

\subsection{Effect of Pseudo-Node Route Augmentation}
\label{sec:exp_route_augmentation}

This experiment evaluates whether pseudo-node augmentation improves the spatial coverage of the high-level route. The routing experiments are performed in \(\Omega=[0,1000]\times[0,1000]\;\mathrm{m}^2\) with randomly generated reported ROI centers. The UAV has a total flight distance budget of \(D_m=5000\;\mathrm{m}\). We compare three routing strategies: an ROI-only route, which visits only the reported ROI centers and the depot; node-based CVT augmentation, which improves spatial waypoint distribution without explicitly considering the current route edges; and edge-based CVT augmentation, which biases pseudo-node placement toward regions under-covered by the current route edges.

Route coverage is evaluated using the edge coverage ratio (ECR) and edge density variance (EDV). The ECR is defined as
\begin{equation}
    \mathrm{ECR}
    =
    \frac{N_c}{N_g},
    \label{eq:metric_ecr}
\end{equation}
where \(N_c\) is the number of grid cells intersected by at least one route edge and \(N_g\) is the total number of grid cells. The EDV is defined as
\begin{equation}
    \mathrm{EDV}
    =
    \frac{1}{N_g}
    \sum_{i=1}^{N_g}
    (X_i-\bar{X})^2,
    \qquad
    \bar{X}
    =
    \frac{1}{N_g}
    \sum_{i=1}^{N_g}X_i,
    \label{eq:metric_edv}
\end{equation}
where \(X_i\) is the number of route edges passing through grid cell \(i\). ECR measures the fraction of the environment intersected by the route, while EDV characterizes how unevenly route-edge density is distributed across the grid.

\begin{figure*}[htbp]
  \centering
  \includegraphics[width=\textwidth, trim=0 0 0 35, clip]{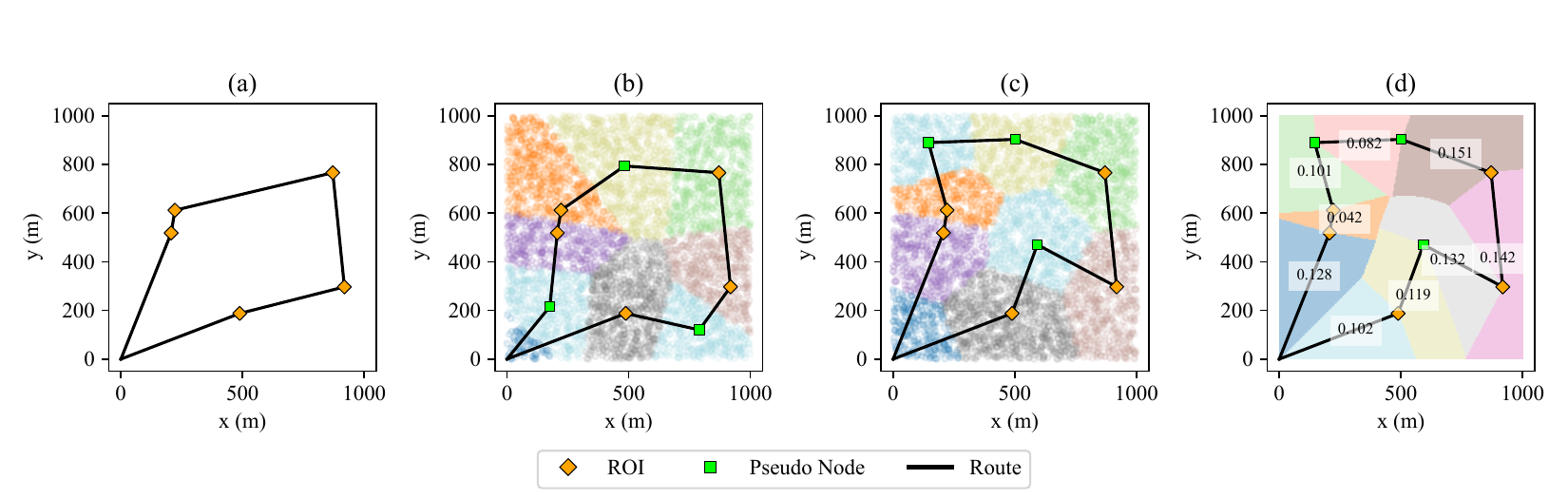}
    \caption{Example of high-level routing strategies. 
    (a) ROI-only routing visits only the reported ROI centers. 
    (b) Node-based CVT introduces pseudo-nodes to improve spatial waypoint distribution. 
    (c) Edge-based CVT biases pseudo-nodes toward regions farther from the existing route edges, improving route-level coverage of under-covered areas. 
    (d) Edge-wise allocation of the remaining flight-distance budget assigns additional exploration allowance to route segments for downstream path planning.}  \label{fig:routing_strategy_sim}
\end{figure*}

\Cref{fig:routing_strategy_sim} illustrates a representative routing example. The ROI-only route visits the required ROI centers but may leave large portions of the domain under-covered. Node-based CVT improves the spatial spread of auxiliary waypoints, while edge-based CVT explicitly accounts for the current route structure and places pseudo-nodes toward regions farther from existing route edges. The resulting augmented route provides more spatially distributed route segments for downstream informative path planning.

The quantitative comparison is summarized in \Cref{tab:cvt_comparison}. The results are averaged over 100 randomized trials. The edge-based CVT strategy achieves the highest ECR, increasing route coverage from \(0.0992\) for the ROI-only route to \(0.1901\). This indicates that the augmented route intersects a substantially larger fraction of the domain. The EDV values increase as pseudo-nodes are added, reflecting a larger number of route-edge intersections across the grid.

\begin{table}[htbp]
\caption{Route Coverage and Edge-Density Metrics for Pseudo-Node Placement Strategies}
\label{tab:cvt_comparison}
\centering
\begin{tabular}{lcc}
\hline
\textbf{Method} & \textbf{ECR} & \textbf{EDV} \\
\hline
ROI-only route & \(0.0992 \pm 0.0245\) & \(0.1177 \pm 0.0305\) \\
Node-based CVT & \(0.1246 \pm 0.0244\) & \(0.1472 \pm 0.0302\) \\
Edge-based CVT & \(0.1901 \pm 0.0358\) & \(0.2367 \pm 0.0495\) \\
\hline
\end{tabular}
\end{table}

These results show that pseudo-node augmentation increases route coverage compared with the ROI-only route. In particular, edge-based CVT provides the largest ECR by biasing pseudo-node placement toward regions that are farther from existing route edges. This broader route-level coverage is beneficial for exploration since the low-level planner receives route segments that span a larger portion of the environment.

\subsection{Representative Online Replanning Behavior}
\label{sec:online_replanning_analysis}

This experiment evaluates the online belief update and replanning component under a representative nominal scenario with \(n_c=8\), \(n_h=3\), and \(N_p=3\). We compare two variants of the proposed framework: an offline optimized planner and an online replanning planner. In both variants, the risk-aware B-spline trajectory optimization problem is solved using IPOPT, and both share the same high-level route, pseudo-node augmentation, edge-wise budget allocation, and initial B-spline trajectory. The only difference is that the online variant updates the belief field during execution and replans the remaining trajectory when the belief-change and budget conditions are satisfied.

Belief-estimation performance is measured using the KL reduction metric introduced in \Cref{eq:kl_reduction}. In the grid-based implementation, \(D_{\mathrm{KL}}(p_{\mathrm{true}}\|b)\) is computed as the average cell-wise Bernoulli KL divergence over \(\mathcal{G}\), with both \(p_{\mathrm{true}}\) and \(b\) clipped to \([\varepsilon,1-\varepsilon]\) to avoid numerical singularities. Larger \(\Delta_{\mathrm{KL}}\) values indicate greater improvement in the UAV's belief estimate.

\Cref{tab:online_replanning_performance} summarizes the offline and online replanning performance for the representative nominal scenario with \(n_c=8\), \(n_h=3\), and \(N_p=3\). Online replanning improves the mean KL reduction from \(0.4953\) to \(0.6037\), corresponding to a \(21.9\%\) improvement over the offline optimized planner. This result shows that updating the belief field during execution and adapting the remaining trajectory can provide a substantial improvement over executing the initially optimized trajectory without replanning.

An important observation is that the improvement is consistent rather than being driven by a small number of favorable trials. In the 100 trial online replanning experiment, the online planner improves KL reduction in every trial, with no negative improvement cases. This suggests that the online update and replanning mechanism is robust to variations in the generated risk fields, rather than only benefiting particular random instances.

\begin{table}[htbp]
\caption{Offline and Online Replanning Performance for the Representative Scenario \(n_c=8\), \(n_h=3\), and \(N_p=3\)}
\label{tab:online_replanning_performance}
\centering
\begin{tabular}{lcc}
\hline
\textbf{Metric} & \textbf{Offline} & \textbf{Online} \\
\hline
\(\Delta_{\mathrm{KL}}\) 
& \(0.4953 \pm 0.0666\) 
& \({0.6037 \pm 0.0757}\) \\
\hline
\end{tabular}
\end{table}

The improvement achieved by online replanning is driven by the interaction between hidden risk clusters, belief-change triggering, and EIG-based replan acceptance. The offline optimized planner allocates sensing effort using the belief field available before execution. Since this initial belief is constructed from reported ROIs, it does not explicitly represent hidden clusters. As the UAV collects measurements during execution, observations near hidden clusters can induce sufficiently large belief updates to exceed the belief-change threshold \(\varepsilon_b\), triggering a replanning evaluation.

\begin{table}[htbp]
\caption{Online Replanning Activity and Per-Call Runtime}
\label{tab:online_replanning_activity_runtime}
\centering
\begin{tabular}{lc}
\hline
\textbf{Metric} & \textbf{Value} \\
\hline
Triggered replanning evaluations per mission & \(46.3 \pm 23.7\) \\
Accepted replans per mission & \(18.4 \pm 9.2\) \\
Mean IPOPT replan call time & \(1.11 \pm 1.11\;\mathrm{s}\) \\
Total IPOPT replan time per mission & \(39.35 \pm 22.73\;\mathrm{s}\) \\
\hline
\end{tabular}
\end{table}

A triggered replanning evaluation does not necessarily replace the current trajectory. Instead, the candidate trajectory is accepted only when its expected information gain exceeds that of the remaining portion of the current trajectory. This EIG-based acceptance criterion makes online replanning selective rather than purely reactive. For the same representative scenario, \Cref{tab:online_replanning_activity_runtime} shows that an average of \(46.3\) replanning evaluations are triggered per mission, but only \(18.4\) are accepted. This indicates that many triggered replanning opportunities are filtered out, preventing unnecessary trajectory changes while still allowing the UAV to redirect sensing effort when new measurements reveal informative regions.

The trigger and acceptance statistics also show that the online planner does not rely on continuous re-optimization. On average, fewer than half of the triggered replanning evaluations are accepted. This behavior is desirable since a low threshold alone could cause frequent reactive replanning, while the EIG-based acceptance test filters candidate trajectories that do not offer enough additional value. Therefore, the online layer acts as a selective adaptation mechanism: it reacts to meaningful belief changes, but only modifies the trajectory when the proposed path is expected to improve information gathering.

From a computational perspective, the cost of online replanning is better characterized by the per-call optimization runtime than by the total simulation wall time. Total wall time can include execution-loop overhead, belief updates, logging, and implementation-dependent costs, whereas the per-call IPOPT runtime directly reflects the cost of solving the trajectory optimization problem. Across the 100 trials in the representative online replanning experiment, the mean IPOPT replan call time was \(1.11 \pm 1.11\) s, with a median of \(0.81\) s. The total IPOPT replan time per mission was \(39.35 \pm 22.73\) s on average.

Although online replanning introduces additional optimization calls during execution, these calls are not made at every time step. The belief-change trigger identifies candidate replanning opportunities, and the EIG-based acceptance criterion further filters them before replacing the current trajectory. The median runtime below one second indicates that most replanning calls are computationally light, while the larger standard deviation and worst-case runtime reflect occasional harder nonlinear trajectory-optimization instances depending on the current state, remaining budget, and local belief structure. Overall, the runtime statistics support the use of online replanning as a tractable adaptation layer rather than a continuously running global optimizer. All IPOPT replanning calls in the reported experiments returned feasible candidate trajectories, and candidates that did not improve the expected information-gain score were rejected by the replanning acceptance test. These results suggest that the online replanning layer is computationally tractable for receding-horizon mission adaptation, although embedded real-time implementation would require further optimization and hardware-specific validation.

\subsection{Path Planning Performance Across Scenarios}
\label{sec:path_planning_performance}

\begin{figure*}[htbp]
  \centering
  \includegraphics[width=\textwidth, trim=0 0 0 0, clip]{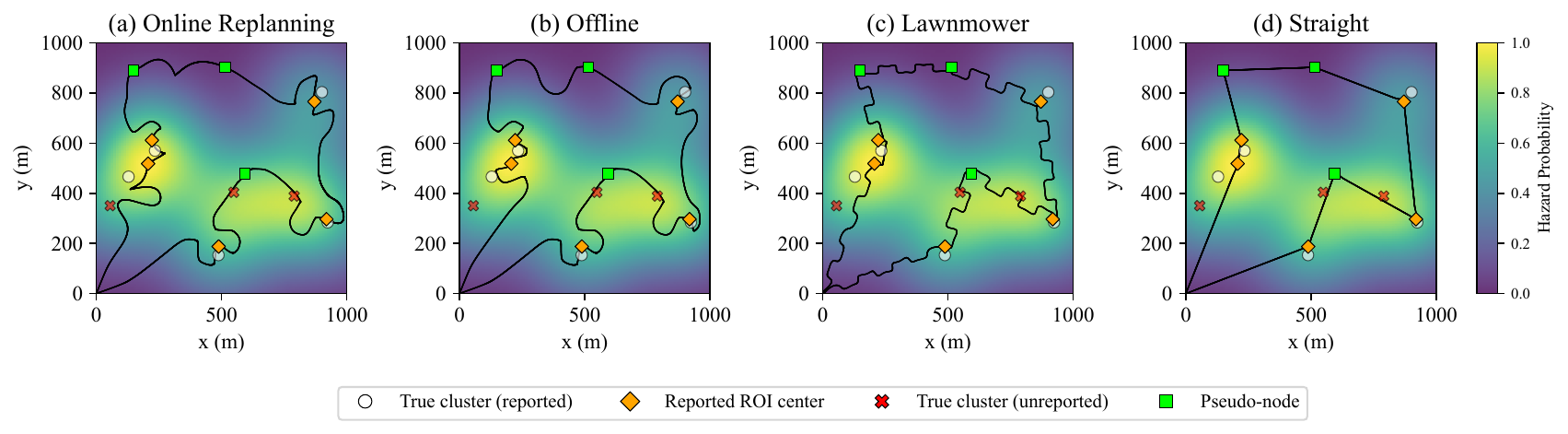}
    \caption{Example trajectories for four path planning strategies. The background shows the true hazard probability field \(p_{\mathrm{true}}\), and the black line indicates the UAV trajectory. Orange diamonds indicate reported ROI centers, and red crosses indicate true hidden cluster centers. (a) Online replanning, (b) offline optimized, (c) lawnmower, and (d) straight-line paths.}
    \label{fig:planned_paths}
\end{figure*}

\begin{table*}[t]
\centering
\scriptsize
\caption{Path planning comparison across 48 scenarios with mean $\pm$ standard deviation over 100 trials. The best-performing method in each scenario is shown in bold.}
\label{tab:path_planning_full_results}
\begin{tabular}{ccccccc}
\toprule
$n_c$ & $n_h$ & $N_p$ & Online & Offline & Lawnmower & Straight \\
\midrule
6 & 1 & 0 & \textbf{0.6556 $\pm$ 0.0736} & 0.5569 $\pm$ 0.0745 & 0.4492 $\pm$ 0.0669 & 0.4128 $\pm$ 0.0717 \\
6 & 1 & 1 & \textbf{0.6162 $\pm$ 0.0628} & 0.5343 $\pm$ 0.0812 & 0.4513 $\pm$ 0.0630 & 0.4113 $\pm$ 0.0738 \\
6 & 1 & 2 & \textbf{0.6386 $\pm$ 0.0640} & 0.5357 $\pm$ 0.0825 & 0.4562 $\pm$ 0.0653 & 0.4160 $\pm$ 0.0687 \\
6 & 1 & 3 & \textbf{0.6280 $\pm$ 0.0601} & 0.5331 $\pm$ 0.0633 & 0.4797 $\pm$ 0.0621 & 0.4397 $\pm$ 0.0564 \\
\cmidrule(l){2-7}
6 & 2 & 0 & \textbf{0.6552 $\pm$ 0.0734} & 0.5630 $\pm$ 0.0857 & 0.4470 $\pm$ 0.0743 & 0.3818 $\pm$ 0.1138 \\
6 & 2 & 1 & \textbf{0.6428 $\pm$ 0.0694} & 0.5417 $\pm$ 0.0699 & 0.4431 $\pm$ 0.0582 & 0.3864 $\pm$ 0.0882 \\
6 & 2 & 2 & \textbf{0.6233 $\pm$ 0.0713} & 0.5411 $\pm$ 0.0824 & 0.4454 $\pm$ 0.0625 & 0.4112 $\pm$ 0.0764 \\
6 & 2 & 3 & \textbf{0.6039 $\pm$ 0.0753} & 0.5325 $\pm$ 0.0600 & 0.4585 $\pm$ 0.0585 & 0.4125 $\pm$ 0.0740 \\
\cmidrule(l){2-7}
6 & 3 & 0 & \textbf{0.5851 $\pm$ 0.0771} & 0.4862 $\pm$ 0.1121 & 0.3775 $\pm$ 0.0965 & 0.3287 $\pm$ 0.1363 \\
6 & 3 & 1 & \textbf{0.5719 $\pm$ 0.0647} & 0.5022 $\pm$ 0.0941 & 0.4065 $\pm$ 0.0896 & 0.3457 $\pm$ 0.0921 \\
6 & 3 & 2 & \textbf{0.6029 $\pm$ 0.0612} & 0.5196 $\pm$ 0.0726 & 0.4233 $\pm$ 0.0711 & 0.3742 $\pm$ 0.0754 \\
6 & 3 & 3 & \textbf{0.6067 $\pm$ 0.0656} & 0.5445 $\pm$ 0.0684 & 0.4510 $\pm$ 0.0640 & 0.4058 $\pm$ 0.0739 \\
\cmidrule(l){2-7}
6 & 4 & 0 & \textbf{0.4711 $\pm$ 0.0605} & 0.4046 $\pm$ 0.1893 & 0.2467 $\pm$ 0.1464 & 0.2532 $\pm$ 0.1310 \\
6 & 4 & 1 & \textbf{0.5383 $\pm$ 0.0618} & 0.4545 $\pm$ 0.1471 & 0.3333 $\pm$ 0.1123 & 0.3129 $\pm$ 0.1304 \\
6 & 4 & 2 & \textbf{0.5683 $\pm$ 0.0744} & 0.4970 $\pm$ 0.1268 & 0.3838 $\pm$ 0.1044 & 0.3484 $\pm$ 0.1159 \\
6 & 4 & 3 & \textbf{0.6213 $\pm$ 0.0692} & 0.5278 $\pm$ 0.0916 & 0.4060 $\pm$ 0.0697 & 0.3670 $\pm$ 0.0817 \\
\midrule
8 & 1 & 0 & \textbf{0.5940 $\pm$ 0.0757} & 0.5293 $\pm$ 0.0743 & 0.4574 $\pm$ 0.0699 & 0.4154 $\pm$ 0.0767 \\
8 & 1 & 1 & \textbf{0.6416 $\pm$ 0.0733} & 0.5328 $\pm$ 0.0593 & 0.4584 $\pm$ 0.0513 & 0.4206 $\pm$ 0.0544 \\
8 & 1 & 2 & \textbf{0.6312 $\pm$ 0.0741} & 0.5331 $\pm$ 0.0673 & 0.4842 $\pm$ 0.0854 & 0.4437 $\pm$ 0.0857 \\
8 & 1 & 3 & \textbf{0.6371 $\pm$ 0.0756} & 0.5378 $\pm$ 0.0678 & 0.4985 $\pm$ 0.0931 & 0.4632 $\pm$ 0.0925 \\
\cmidrule(l){2-7}
8 & 2 & 0 & \textbf{0.6141 $\pm$ 0.0727} & 0.5264 $\pm$ 0.0801 & 0.4438 $\pm$ 0.0662 & 0.4257 $\pm$ 0.0818 \\
8 & 2 & 1 & \textbf{0.5909 $\pm$ 0.0711} & 0.5277 $\pm$ 0.0719 & 0.4494 $\pm$ 0.0632 & 0.3978 $\pm$ 0.0661 \\
8 & 2 & 2 & \textbf{0.6256 $\pm$ 0.0712} & 0.5242 $\pm$ 0.0645 & 0.4528 $\pm$ 0.0642 & 0.4137 $\pm$ 0.0633 \\
8 & 2 & 3 & \textbf{0.6117 $\pm$ 0.0661} & 0.5201 $\pm$ 0.0628 & 0.4639 $\pm$ 0.0550 & 0.4237 $\pm$ 0.0535 \\
\cmidrule(l){2-7}
8 & 3 & 0 & \textbf{0.5579 $\pm$ 0.0659} & 0.4902 $\pm$ 0.0779 & 0.3963 $\pm$ 0.0807 & 0.3650 $\pm$ 0.0936 \\
8 & 3 & 1 & \textbf{0.6109 $\pm$ 0.0732} & 0.5167 $\pm$ 0.0848 & 0.4069 $\pm$ 0.0700 & 0.3709 $\pm$ 0.0744 \\
8 & 3 & 2 & \textbf{0.6102 $\pm$ 0.0711} & 0.5130 $\pm$ 0.0682 & 0.4318 $\pm$ 0.0525 & 0.3960 $\pm$ 0.0616 \\
8 & 3 & 3 & \textbf{0.6037 $\pm$ 0.0757} & 0.4953 $\pm$ 0.0666 & 0.4313 $\pm$ 0.0635 & 0.3977 $\pm$ 0.0658 \\
\cmidrule(l){2-7}
8 & 4 & 0 & \textbf{0.5619 $\pm$ 0.0632} & 0.4800 $\pm$ 0.0895 & 0.3771 $\pm$ 0.0899 & 0.3481 $\pm$ 0.0945 \\
8 & 4 & 1 & \textbf{0.6036 $\pm$ 0.0700} & 0.5020 $\pm$ 0.0823 & 0.3882 $\pm$ 0.0772 & 0.3582 $\pm$ 0.0743 \\
8 & 4 & 2 & \textbf{0.6171 $\pm$ 0.0630} & 0.5221 $\pm$ 0.0667 & 0.4183 $\pm$ 0.0647 & 0.3770 $\pm$ 0.0650 \\
8 & 4 & 3 & \textbf{0.6099 $\pm$ 0.0739} & 0.5343 $\pm$ 0.0715 & 0.4474 $\pm$ 0.0613 & 0.4073 $\pm$ 0.0616 \\
\midrule
10 & 1 & 0 & \textbf{0.6030 $\pm$ 0.0692} & 0.5366 $\pm$ 0.0626 & 0.4875 $\pm$ 0.0606 & 0.4582 $\pm$ 0.0655 \\
10 & 1 & 1 & \textbf{0.6171 $\pm$ 0.0714} & 0.5346 $\pm$ 0.0690 & 0.5017 $\pm$ 0.0725 & 0.4790 $\pm$ 0.0734 \\
10 & 1 & 2 & \textbf{0.6117 $\pm$ 0.0628} & 0.5332 $\pm$ 0.0630 & 0.5030 $\pm$ 0.0648 & 0.4730 $\pm$ 0.0637 \\
10 & 1 & 3 & \textbf{0.6612 $\pm$ 0.0623} & 0.5549 $\pm$ 0.0842 & 0.5334 $\pm$ 0.0979 & 0.5060 $\pm$ 0.0948 \\
\cmidrule(l){2-7}
10 & 2 & 0 & \textbf{0.6112 $\pm$ 0.0606} & 0.5133 $\pm$ 0.0610 & 0.4570 $\pm$ 0.0490 & 0.4180 $\pm$ 0.0638 \\
10 & 2 & 1 & \textbf{0.5933 $\pm$ 0.0687} & 0.5156 $\pm$ 0.0521 & 0.4566 $\pm$ 0.0457 & 0.4195 $\pm$ 0.0497 \\
10 & 2 & 2 & \textbf{0.6276 $\pm$ 0.0643} & 0.5191 $\pm$ 0.0542 & 0.4672 $\pm$ 0.0483 & 0.4355 $\pm$ 0.0481 \\
10 & 2 & 3 & \textbf{0.6333 $\pm$ 0.0682} & 0.5286 $\pm$ 0.0517 & 0.4878 $\pm$ 0.0552 & 0.4610 $\pm$ 0.0571 \\
\cmidrule(l){2-7}
10 & 3 & 0 & \textbf{0.6012 $\pm$ 0.0733} & 0.5249 $\pm$ 0.0869 & 0.4490 $\pm$ 0.0644 & 0.4330 $\pm$ 0.0570 \\
10 & 3 & 1 & \textbf{0.6046 $\pm$ 0.0681} & 0.5261 $\pm$ 0.0754 & 0.4552 $\pm$ 0.0500 & 0.4108 $\pm$ 0.0630 \\
10 & 3 & 2 & \textbf{0.6119 $\pm$ 0.0763} & 0.5284 $\pm$ 0.0578 & 0.4746 $\pm$ 0.0505 & 0.4402 $\pm$ 0.0591 \\
10 & 3 & 3 & \textbf{0.6039 $\pm$ 0.0633} & 0.5344 $\pm$ 0.0667 & 0.4938 $\pm$ 0.0762 & 0.4608 $\pm$ 0.0767 \\
\cmidrule(l){2-7}
10 & 4 & 0 & \textbf{0.6512 $\pm$ 0.0689} & 0.5496 $\pm$ 0.0983 & 0.4476 $\pm$ 0.0837 & 0.4082 $\pm$ 0.0862 \\
10 & 4 & 1 & \textbf{0.6104 $\pm$ 0.0676} & 0.5102 $\pm$ 0.0794 & 0.4585 $\pm$ 0.0773 & 0.4245 $\pm$ 0.0816 \\
10 & 4 & 2 & \textbf{0.5970 $\pm$ 0.0660} & 0.5176 $\pm$ 0.0722 & 0.4623 $\pm$ 0.0660 & 0.4228 $\pm$ 0.0675 \\
10 & 4 & 3 & \textbf{0.6158 $\pm$ 0.0726} & 0.5414 $\pm$ 0.0739 & 0.4767 $\pm$ 0.0817 & 0.4419 $\pm$ 0.0859 \\
\bottomrule
\end{tabular}
\end{table*}

\begin{figure}[ht]
    \centering
    \includegraphics[width=0.85\linewidth]{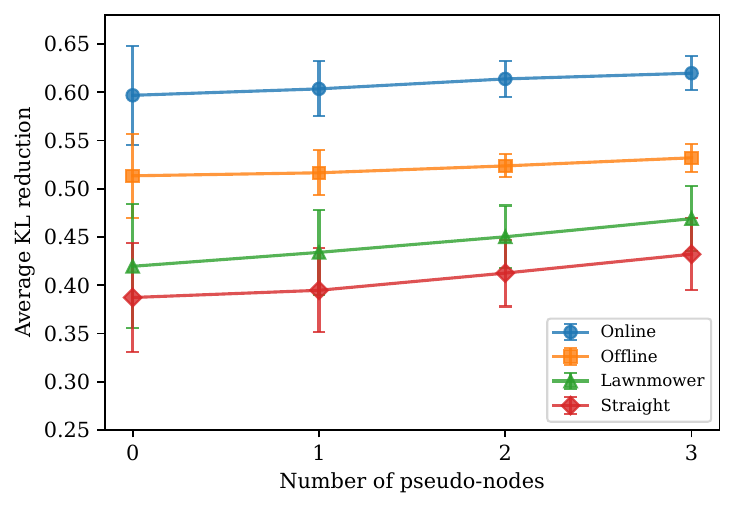}
    \caption{Average KL reduction as a function of the number of pseudo-nodes \(N_p\), averaged across all scenarios. Error bars indicate standard deviations across scenarios. Increasing \(N_p\) generally improves performance, while the proposed online replanning method consistently achieves the highest KL reduction.}
    \label{fig:pseudo_nodes_overall_average}
\end{figure}

This experiment evaluates path planning performance after the high-level route has been computed. For each route edge, we compare four execution strategies: straight-line traversal, lawnmower coverage~\cite{GALCERAN20131258}, the proposed offline optimized B-spline planner, and the proposed online replanning planner. The straight-line baseline directly connects consecutive route waypoints and therefore represents a minimum-travel strategy with limited exploration and no belief-field use. The lawnmower baseline uses the assigned path budget to sweep a local region around each route edge, representing a non-adaptive geometric coverage strategy that uses the marginal exploration budget but does not use the belief field. The proposed offline optimized planner solves the risk-aware B-spline trajectory optimization problem using the belief field available before execution, but then executes the resulting trajectory without replanning. The proposed online replanning planner starts from the offline optimized trajectory, updates the belief field using streaming measurements, and replans the remaining trajectory when the belief-change and budget conditions are satisfied.

This comparison separates the effects of marginal budget use, geometric exploration, belief-aware trajectory optimization, and online adaptation. The straight-line baseline evaluates direct route traversal with minimal use of the marginal exploration budget. The lawnmower baseline evaluates whether using the assigned path budget for additional geometric coverage improves belief-field estimation without belief-aware planning. The offline optimized planner evaluates the benefit of belief-aware B-spline trajectory optimization under the initial belief field. The online replanning planner evaluates the additional benefit of adapting the remaining trajectory as new measurements are collected. All path planning comparisons use the same KL reduction metric and grid-based Bernoulli KL implementation described above. 

\Cref{fig:planned_paths} illustrates representative trajectories of the four methods on the same environment instance. Orange diamonds mark reported ROI centers used as routing nodes, and red crosses indicate true cluster centers hidden from the planner. The online replanning method visibly deviates from the nominal offline trajectory near hidden cluster regions, redirecting the remaining path as streaming measurements update the belief field. In contrast, the offline optimized, lawnmower, and straight-line methods follow fixed trajectories that do not respond to in-flight measurements.

To evaluate robustness across different environment structures, we vary the total number of clusters \(n_c \in \{6,8,10\}\), the number of hidden clusters \(n_h \in \{1,2,3,4\}\), and the number of pseudo-nodes \(N_p \in \{0,1,2,3\}\), as summarized in \Cref{tab:simulation_parameters}. This gives \(3 \times 4 \times 4 = 48\) scenario configurations. \Cref{tab:path_planning_full_results} reports the KL reduction for all four methods across these scenarios, where Online and Offline denote the proposed online replanning planner and the proposed offline optimized B-spline planner, respectively, and Lawnmower and Straight denote the two non-belief-aware baselines. For each scenario, the same generated risk fields, reported ROIs, routes, and flight distance budgets are used across all methods to ensure a fair comparison.

The results in \Cref{tab:path_planning_full_results} show a clear ordering of performance across the four methods. Averaged over all 48 scenario configurations, the online replanning planner achieves the largest mean KL reduction, \(0.6043\), followed by the offline optimized planner with \(0.5214\), the lawnmower baseline with \(0.4433\), and the straight-line baseline with \(0.4068\). Thus, online replanning improves the average KL reduction by \(15.9\%\) relative to the offline optimized planner, \(36.3\%\) relative to lawnmower coverage, and \(48.6\%\) relative to straight-line traversal. The offline optimized planner also improves over the straight-line and lawnmower baselines, showing that belief-aware trajectory optimization provides a substantial benefit beyond geometric coverage alone.

The ordering of the four methods reflects the incremental use of planning capabilities and available mission resources. Straight-line traversal provides the minimum-travel baseline and uses little of the marginal exploration budget. Lawnmower coverage improves over straight-line traversal by using the assigned budget for additional geometric coverage, but it does not exploit the belief field. The offline optimized planner further improves performance by using both the marginal path budget and the initial belief field to allocate sensing effort toward informative regions. The online replanning planner achieves the best performance since it additionally incorporates streaming measurements and redirects the remaining trajectory when the updated belief reveals mislocalized or previously unreported risk. 

The lawnmower baseline provides a useful intermediate comparison since it isolates the value of geometric exploration from belief-aware planning. Its improvement over the straight-line path shows that using the marginal path budget for additional spatial coverage is beneficial. However, the gap between lawnmower coverage and the offline optimized planner shows that coverage alone is not sufficient. Since the lawnmower pattern is not conditioned on the belief field, it can spend sensing effort in regions that are less informative for risk field estimation. In contrast, the optimized B-spline planner uses the current belief to bias the trajectory toward higher-value sensing regions, which explains its consistent improvement over the non-belief-aware baselines.

The scenario-level trends across \(n_h\), \(n_c\), and \(N_p\) provide additional insight into when each component is most useful. Increasing \(n_h\) makes the prior belief less complete as a larger portion of the true risk field is not represented by reported ROIs. This setting is particularly challenging for the offline optimized planner, which can only exploit the initial belief. Online replanning is better suited to this case since measurements collected during execution can reveal discrepancies between the prior and the true field, allowing the remaining trajectory to be redirected. Increasing \(N_p\) generally improves the performance of both offline and online planning since pseudo-nodes create route segments that span a broader portion of the environment, giving the low-level planner more opportunities to use the exploration budget. The effect of \(n_c\) is more nuanced: as the number of total clusters increases, risk becomes more spatially distributed, which can also benefit geometric baselines, but the online planner remains the strongest method as it can still adapt to local belief updates.

Overall, these results demonstrate that the proposed framework improves monitoring performance through complementary route-level, trajectory-level, and execution-time mechanisms. Pseudo-node augmentation improves route-level coverage, offline B-spline optimization improves sensing efficiency by using the prior belief field, and online replanning further improves belief accuracy by adapting the remaining trajectory as new measurements are collected.

\section{Conclusion and Future Work}
\label{sec:conclusion}

This paper presented an integrated routing and online informative path planning framework for UAV hazard monitoring under imperfect and evolving information. Reported high-risk areas were modeled as uncertain ROIs rather than confirmed hazard locations, enabling the UAV to satisfy required inspection tasks while estimating a broader continuous risk field. The proposed framework combines ROI-constrained routing, pseudo-node route augmentation, edge-wise exploration budget allocation, risk-aware B-spline trajectory optimization, and online belief-driven replanning.

Numerical experiments demonstrated the effectiveness of the proposed approach. Pseudo-node augmentation improved route-level spatial coverage compared with ROI-only routing, while the offline informative planner outperformed straight-line and lawnmower baselines by exploiting the prior belief field for sensing-aware trajectory generation. Online replanning further improved estimation performance by adapting the remaining trajectory as streaming measurements updated the belief field. Across 48 scenario configurations, online replanning achieved a 15.9\% improvement in average KL reduction relative to the offline optimized planner, showing the value of online adaptation when prior hazard information is incomplete or mislocalized.

Several limitations remain. First, the numerical experiments used synthetic, fixed ground-truth risk fields to enable controlled comparison across planning methods. Although this setup is useful for isolating the effects of routing, informative planning, and online replanning, it does not fully capture time-varying hazards, complex terrain, or real sensor artifacts. Second, the current numerical experiments focuse on a single UAV setting with a centralized belief representation and does not address communication constraints, decentralized coordination, or conflict resolution among multiple UAVs. Third, pseudo-node selection and edge-wise budget allocation are treated as mission-design components rather than fully adaptive decisions during execution. Finally, the sensor model is intentionally simplified and should be validated against higher-fidelity sensing models or real UAV monitoring data.

Future work will address these limitations by extending the framework to explicitly time-varying risk fields, adaptive pseudo-node selection, online budget reallocation, and coordinated multi-UAV missions with shared or distributed belief updates. We also plan to evaluate the approach with higher-fidelity sensing models, more realistic vehicle and environmental constraints, and real-world UAV monitoring data.





\section*{Conflict of Interest}
The authors declare no conflict of interest.

\bibliographystyle{IEEEtran}
\bibliography{reference}  

\end{document}